\documentclass{article}

\PassOptionsToPackage{numbers, compress}{natbib}

\usepackage[final]{neurips_2024}

\usepackage[utf8]{inputenc}
\usepackage[T1]{fontenc}
\usepackage[table,xcdraw]{xcolor}
\usepackage{booktabs}
\usepackage{nicefrac}
\usepackage{microtype}
\usepackage{mathtools}
\usepackage{amsthm}
\usepackage{cancel}
\usepackage{algpseudocode}
\usepackage{algorithm}
\usepackage{xifthen}
\usepackage{amsfonts}
\usepackage{multirow}
\usepackage{subcaption}
\usepackage[acronym]{glossaries}
\usepackage{hyperref}
\usepackage{cleveref}
\usepackage{wrapfig}

\def\N{\mathcal{N}}
\def\E{\mathbb{E}}
\def\R{\mathbb{R}}
\def\O{\mathcal{O}}
\def\x{\mathbf{x}}
\def\hx{\hat{\x}}
\def\z{\mathbf{z}}
\def\bz{\bar{\z}}
\def\bq{\bar{q}}
\def\bp{\bar{p}}
\def\w{\mathbf{w}}
\def\bw{\bar{\w}}
\def\a{\alpha}
\def\s{\sigma}
\def\bmu{\bar{\mu}}
\def\bs{\bar{\s}}
\def\l{\lambda}
\def\d{\delta}
\def\e{\varepsilon}
\def\ph{\varphi}
\def\th{\theta}
\def\tff{\tilde{f}^F}
\def\tfb{\tilde{f}^B}
\def\hf{\hat{f}}
\def\D{\Delta}
\def\p{\partial}
\def\n{\nabla}
\def\bF{\bar{F}}
\def\KL{{\mathrm{D}_\mathrm{KL}}}
\def\L{\mathcal{L}}
\def\Lrec{{\L_\mathrm{rec}}}
\def\Ldiff{{\L_\mathrm{diff}}}
\def\Lprior{{\L_\mathrm{prior}}}
\def\Lot{{\L_\mathrm{OT}}}
\def\Lcrv{{\L_\mathrm{crv}}}
\def\Lsimple{{\L_\mathrm{simple}}}
\def\bL{\bar{\L}}
\def\bLrec{{\bL_\mathrm{rec}}}
\def\bLdiff{{\bL_\mathrm{diff}}}
\def\bLprior{{\bL_\mathrm{prior}}}

\newcommand{\TODO}[1][]{{\ifthenelse{\isempty{#1}}{\color{red}\bf TODO}{\color{red}\bf (TODO: #1)}}}

\theoremstyle{plain}

\theoremstyle{definition}

\theoremstyle{remark}

\makeglossaries
\newglossaryentry{fproc}{
    name={forward process},
    plural={forward processes},
    description={}
}
\newglossaryentry{rproc}{
    name={reverse process},
    plural={reverse processes},
    description={}
}
\newglossaryentry{simfree}{
    name={simulation-free},
    description={}
}
\newacronym{nfdm}{NFDM}{Neural Flow Diffusion Models}
\newacronym{nfbm}{NFBM}{Neural Flow Bridge Models}
\newacronym{sde}{SDE}{Stochastic Differential Equation}
\newacronym{ode}{ODE}{Ordinary Differential Equation}
\newacronym{elbo}{ELBO}{Evidence Lower Bound}
\newacronym{nll}{NLL}{negative log-likelihood}
\newacronym{bpd}{BPD}{Bits Per Dimension}
\newacronym{fid}{FID}{Frechet Inception Distance}
\newacronym{nfe}{NFE}{Number of Function Evaluations}
\newacronym{kl}{KL}{Kullback–Leibler}
\newacronym{cdf}{CDF}{Cumulative Distribution Function}
\newacronym{vjp}{VJP}{Vector Jacobian Product}
\newacronym{jvp}{JVP}{Jacobian Vector Product}
\newacronym{vae}{VAE}{Variational Autoencoder}
\newacronym{cnf}{CNF}{Continuous Normalizing Flow}

\title{Neural Flow Diffusion Models: Learnable Forward Process for Improved Diffusion Modelling}

\author{
  Grigory Bartosh \\
  University of Amsterdam\\
  \texttt{g.bartosh@uva.nl} \\
  \And
  Dmitry Vetrov \\
  Constructor University, Bremen\\
  \texttt{dvetrov@constructor.university} \\
  \And
  Christian A. Naesseth \\
  University of Amsterdam\\
  \texttt{c.a.naesseth@uva.nl} \\
}

\begin{document}

\maketitle

\begin{abstract}
  Conventional diffusion models often rely on a fixed \gls{fproc}, which implicitly defines complex marginal distributions over latent variables. This can often complicate the \gls{rproc}' task in learning generative trajectories, and results in costly inference for diffusion models. To address these limitations, we introduce \gls{nfdm}, a novel framework that enhances diffusion models by supporting a broader range of forward processes beyond the standard linear Gaussian. We also propose a novel parameterization technique for learning the \gls{fproc}. Our framework provides an end-to-end, simulation-free optimization objective, effectively minimizing a variational upper bound on the \acrlong{nll}. Experimental results demonstrate \gls{nfdm}'s strong performance, evidenced by state-of-the-art likelihoods across a range of image generation tasks. Furthermore, we investigate \gls{nfdm}'s capacity for learning generative dynamics with specific characteristics, such as deterministic straight lines trajectories, and demonstrate how the framework can be adopted for learning bridges between two distributions. The results underscores \gls{nfdm}'s versatility and its potential for a wide range of applications.
\end{abstract}

\glsresetall

\section{Introduction}
\label{sec:introduction}


Diffusion models \cite{sohl2015deep, ho2020denoising} are a class of generative models constructed by two key processes: the \gls{fproc} and the \gls{rproc}. The \gls{fproc} gradually corrupts the data distribution, transforming it from its original form to a noised state. The \gls{rproc} learns to invert corruptions of the \gls{fproc} and restore the data distribution. This way, the model learns to generate data from pure noise. Diffusion models have demonstrated remarkable results in various domains \cite{ho2022cascaded, saharia2022photorealistic, tan2024naturalspeech, watson2023novo, trippe2023diffusion}. Nevertheless, most existing diffusion models fix the \gls{fproc} to be predefined, usually linear, Gaussian which makes it unable to adapt to the task at hand or simplify the target for the \gls{rproc}. At the same time there is a growing body of work that demonstrates how modifications of the \gls{fproc} improve performance in terms of generation quality \cite{nichol2021improved, vahdat2021score, daras2022soft}, likelihood estimation \cite{kingma2021variational, nielsen2024diffenc, bartosh2023neural} or sampling speed \cite{lee2023minimizing, pooladian2023multisample, tong2023improving}.

In this paper, we present \gls{nfdm}, a framework that allows for the pre-specification and learning of complex latent variable distributions defined by the \gls{fproc}. Unlike conventional diffusion models \cite{ho2020denoising}, which rely on a conditional Gaussian \gls{fproc}, \gls{nfdm} can accommodate any continuous (and learnable) distribution that can be expressed as an invertible mapping applied to noise. We also derive, and leverage, a new end-to-end \gls{simfree} optimization procedure for \gls{nfdm}, that minimizes a variational upper bound on the \gls{nll}.

Furthermore, we propose an efficient neural network-based parameterization for the \gls{fproc}, enabling it to adapt to the \gls{rproc} during training and simplify the learning of the data distribution. To demonstrate \gls{nfdm}'s capabilities with a learnable \gls{fproc} we provide experimental results on CIFAR-10, ImageNet 32 and 64, attaining state-of-the-art \gls{nll} results, which is crucial for many applications such as data compression \cite{ho2021anfic, yang2024lossy}, anomaly detection \cite{chen2018autoencoder, dias2020anomaly} and  out-of-distribution detection \cite{serra2020input, xiao2020likelihood}.

Then, leveraging the flexibility of \gls{nfdm}, we demonstrate how this framework can be applied to learn bridges between two distributions using the AFHQ dataset. Finally, we explore training with constraints on the \gls{rproc} to learn generative dynamics with specific properties. As a case study, we discuss curvature and obstacle avoidance penalties on the deterministic generative trajectories. Our empirical results indicate improved computational efficiency compared to baselines on CIFAR-10, downsampled ImageNet, and synthetic data.

We summarize our contributions as follows:
\begin{enumerate}
    \item We introduce \acrlong{nfdm} (\gls{nfdm}), improving diffusion modelling through a learnable \gls{fproc}.

    \item We develop an end-to-end optimization procedure that minimizes an upper bound on the \acrlong{nll} in a \gls{simfree} manner.

    \item We demonstrate state-of-the-art log-likelihood results on CIFAR-10, ImageNet 32 and 64.

    \item We show how \gls{nfdm} can be used in learning bridges and generative processes with specific properties, such as dynamics with straight line trajectories, leading to significantly faster sampling speeds and enhanced generation quality with fewer sampling steps.
\end{enumerate}

\section{Background}
\label{sec:background}

Diffusion models are generative latent variable models consisting of two processes: the forward and the reverse (or generative) process. The \gls{fproc} is a dynamic process that takes a data point $\x \sim q(\x), \x \in \R^D$, and perturbs it over time by injecting noise. This generates a trajectory of latent variables $\{\z(t)\}_{t \in [0, 1]}$, conditional on the data $\x$, where $[0, 1]$ is a fixed time horizon and $\z_t = \z(t) \in \R^D$. The (conditional) distribution can be described by an initial distribution $q(\z_0|\x)$ and a \gls{sde} with a linear drift term $\tff(\z_t, t): \R^D \times [0, 1] \mapsto \R^D$, scalar variance $g(t): [0, 1] \mapsto \R_+$, and a standard Wiener process $\w$:
\begin{align}
    d \z_t &= \tff(\z_t, t) d t + g(t) d \w.
    \label{eq:marginal_f_sde}
\end{align}
Due to the linearity of $\tff$, we can reconstruct the conditional marginal distribution $q(\z_t|\x) = \N(\z_t; \a_t \x, \s_t^2 I)$. Typically, the conditional distributions evolve from some low variance distribution $q(\z_0|\x) \approx \d(\z_0 - \x)$ to a unit Gaussian $q(\z_1|\x) \approx \N(\z_1; 0, I)$. This \gls{fproc} is then reversed by starting from the prior $\z_1 \sim \N(\z_1; 0, I)$, and following the reverse \gls{sde} \cite{anderson1982reverse}:
\begin{align}
    d \z_t = \tfb(\z_t, t) d t + g(t) d \bw, 
    \quad \textrm{where} \quad
    \tfb(\z_t, t) = \tff(\z_t, t) - g^2(t) \n_{\z_t} \log q(\z_t).
    \label{eq:marginal_f_rsde}
\end{align}
Here, $\bw$ denotes a standard Wiener process where time flows backwards. Diffusion models approximate this \gls{rproc} by learning $\n_{\z_t} \log q(\z_t)$, known as the score function, through a $\l_t$-weighted denoising score matching loss:
\begin{align}
    \underset{u(t) q(\x, \z_t)}{\E} \left[ \l_t \big\| s_\th(\z_t, t) - \n_{\z_t} \log q(\z_t|\x) \big\|_2^2 \right],
    \label{eq:den_score_matching}
\end{align}
where $u(t)$ represents a uniform distribution over the interval $[0, 1]$, and $s_\th: \R^D \times [0, 1] \mapsto \R^D$ is a learnable approximation. With a learned score function $s_\th(\z_t, t)$, one can generate a sample from the \gls{rproc} by first sampling from the prior $\z_1 \sim \N(\z_1; 0, I)$, and then simulating the reverse \gls{sde}, resulting in a sample $\z_0 \sim p_\th(\z_0) \approx q(\z_0) \approx q(\x)$:
\begin{align}
    d \z_t &= \big[ \tff(\z_t, t) - g^2(t) s_\th(\z_t, t) \big] d t + g(t) d \bw.
    \label{eq:back_r_rsde}
\end{align}

Diffusion models possess several important properties. For example, for a specific $\l_t$, the objective~(\cref{eq:den_score_matching}) can be reformulated \cite{song2021maximum} as an \gls{elbo} on the model's likelihood. Furthermore, the minimization of denoising score matching~(\cref{eq:den_score_matching}) is a \gls{simfree} procedure. This means that simulating either the forward or reverse processes through its \gls{sde} is not necessary for sampling $\z_t$, nor is it necessary for estimating the gradient of the loss function. Instead, we can directly sample $\z_t \sim q(\z_t|\x)$. The \gls{simfree} nature of this approach is a crucial aspect for efficient optimization.

Another notable property is the  existence of an \gls{ode} corresponding to the same marginal densities $q(\z_t)$ as the \gls{sde}~(\cref{eq:marginal_f_sde}):
\begin{align}
    d \z_t &= f(\z_t, t) d t, \quad \textrm{where} \\
    f(\z_t, t) &= \tff(\z_t, t) - \frac{g^2(t)}{2} \n_{\z_t} \log q_\ph(\z_t).
\end{align}

This implies that we can sample from diffusion models deterministically, allowing the use of off-the-shelf numerical \gls{ode} solvers for sampling, which may improve the sampling speed compared to stochastic sampling that requires simulating an \gls{sde}. Additionally, deterministic sampling enables us to compute densities by treating the model as a continuous normalizing flow, as detailed in \cite{chen2018neural, grathwohl2018scalable}.

\section{Neural Flow Diffusion Models}
\label{sec:nfdm}

In diffusion models the \gls{fproc} defines stochastic conditional trajectories $\{\z(t)\}_{t \in [0, 1]}$ and the \gls{rproc} tries to match the marginal distribution of trajectories. This construction can be viewed as a specific type of hierarchical \glspl{vae} \cite{kingma2014auto, rezende2014stochastic}. However, in conventional diffusion models the latent variables are inferred through a pre-specified linear combination of the data point and Gaussian noise. This formulation limits diffusion models in terms of the flexibility of their latent space, and makes learning of the \gls{rproc} more challenging. To address this limitation, we propose a generalized form of the \gls{fproc} that enables the definition and learning of a broad range of distributions in the latent space. From a practical perspective, a more flexible \gls{fproc} can simplify the task of learning the \gls{rproc}. From a theoretical perspective, learning of the \gls{fproc} is analogous to learning the variational distribution in a hierarchical \gls{vae}, which gives a tighter bound on model's \gls{nll} (see an extended discussion in \Cref{app:nfdm_motivation}).

In this section, we introduce \acrfull{nfdm} -- a framework that generalizes conventional diffusion models. The key idea in \gls{nfdm} is to define the \gls{fproc}' conditional \gls{sde} implicitly via a learnable transformation $F_\ph(\e, t, \x)$ that defines the marginal distributions. This lets the user define a broad range of continuous time- and data-dependent \glspl{fproc}, that the \gls{rproc} will learn to invert. Importantly, \gls{nfdm} retains crucial properties of conventional diffusion models, like likelihood-based and \gls{simfree} training. Previous diffusion models emerge as special cases when the data transformation is linear, time-independent, and/or additive Gaussian.

\subsection{Forward Process}
\label{sec:nfdm_forward}

We approach the \gls{fproc} constructively. The ultimate goal is a learnable distribution over trajectories, $\{\z(t)\}_{t \in [0, 1]}$ given $\x$, realized by a conditional \gls{sde} constructed in three steps.

First, we (implicitly) define the conditional marginal distributions $q_\ph(\z_t|\x)$ for $t \in [0,1]$ using $F_\ph(\e, t, \x)$. Then, we introduce the corresponding conditional \gls{ode} that together with an initial distribution $q_\ph(\z_0|\x)$ matches the conditional marginal distribution $q_\ph(\z_t|\x)$. Finally, we define a conditional \gls{sde} that defines a distribution over trajectories $\{\z(t)\}_{t \in [0, 1]}$, with marginal distributions $q_\ph(\z_t|\x)$ for each $t$.

\textbf{Forward Marginal Distribution}. We characterize the marginal distribution $q_\ph(\z_t|\x)$ of the \gls{fproc} trough a function that transforms a noise sample $\e$ into $\z_t$, conditional on the time step $t$ and data point $\x$:
\begin{align}
    \z_t = F_\ph(\e, t, \x),
    \label{eq:f}
\end{align}
where $F_\ph: \R^D \times [0,1] \times \R^D \mapsto \R^D$ and $\e \sim q(\e) = \N(\e; 0, I)$. This defines the conditional distribution of latent variables $q_\ph(\z_t|\x)$. Additionally, \cref{eq:f} facilitates direct and efficient sampling from $q_\ph(\z_t|\x)$ through $F_\ph$ in \cref{eq:f}.

\textbf{Conditional \gls{ode}}. We assume that $F_\ph$ is differentiable with respect to $\e$ and $t$, invertible with respect to $\e$. Further, we assume that fixing specific values of $\x$ and $\e$ and varying $t$ from $0$ to $1$ results in a smooth trajectory from $\z_0$ to $\z_1$. Differentiating these trajectories over time yields a velocity field corresponding to the conditional distribution $q_\ph(\z_t|\x)$, thereby defining a conditional \gls{ode}:
\begin{align}
    d \z_t = f_\ph(\z_t, t, \x) d t, 
    \quad \textrm{where} \quad
    f_\ph(\z_t, t, \x) = \left. \frac{\p F_\ph(\e, t, \x)}{\p t} \right|_{\e=F_\ph^{-1}(\z_t, t, \x)}.
    \label{eq:f_ode}
\end{align}

Therefore, if we sample $\z_0 \sim q_\ph(\z_0|\x)$ and solve \cref{eq:f_ode} until time $t$, we have $\z_t \sim q_\ph(\z_t|\x)$.

The time derivative of $F_\ph$ may be calculated efficiently using automatic differentiation tools like PyTorch \cite{paszke2017automatic} or JAX \cite{jax2018github} (see details in \Cref{app:nfdm_time_dir}).

\textbf{Conditional \gls{sde}}. The function $F_\ph$ and the distribution of the noise $q(\e)$ together defines $q_\ph(\z_t|\x)$. To completely define the distribution of trajectories $\{\z(t)\}_{t \in [0, 1]}$, we introduce a conditional \gls{sde} that starts from sample $\z_0$ and runs forward in time.

With access to both the \gls{ode} and score function $\n_{\z_t} \log q_\ph(\z_t|\x)$, an \gls{sde} \cite{song2021scorebased} with marginal distributions $q_\ph(\z_t|\x)$, is:
\begin{align}
    d \z_t = \tff_\ph(\z_t, t, \x) d t + g_\ph(t) d \w, \quad \textrm{where} 
    \label{eq:f_sde} \\
    \tff_\ph(\z_t, t, \x) = f_\ph(\z_t, t, \x) + \frac{g_\ph^2(t)}{2} \n_{\z_t} \log q_\ph(\z_t|\x). \nonumber
\end{align}
Here, $g_\ph: [0, 1] \mapsto \R_+$ is a scalar function, and $\w$ represents a standard Wiener process. Note that $g_\ph$ only influences the distribution of trajectories $\{\z(t)\}_{t \in [0, 1]}$. The marginal distributions are the same, $q_\ph(\z_t|\x)$, for any choice of $g_\ph$.

The \gls{sde} in \cref{eq:f_sde} requires access to the conditional score function $\n_{\z_t} \log q_\ph(\z_t|\x)$, which in the general case can be computationally costly. However, for $F_\ph$ that allow efficient evaluation of the log-determinant of the Jacobian matrix, the score function can be calculated efficiently. Examples of such $F_\ph$ are functions linear with respect to $\e$ or RealNVP style architectures \cite{dinh2017density, kingma2018glow}. The calculation of this score function is further discussed in \Cref{app:nfdm_score}.

\subsection{Reverse (Generative) Process}
\label{sec:nfdm_reverse}

To define the reverse (generative) process, we specify a reverse \gls{sde} that starts from $\z_1 \sim p(\z_1)$ and runs backwards in time. To do so we first introduce a conditional reverse \gls{sde} that reverses the conditional forward \gls{sde}~(\cref{eq:f_sde}). Following \cite{anderson1982reverse}, we define:
\begin{align}
    d \z_t &= \tfb_\ph(\z_t, t, \x) d t + g_\ph(t) d \bw, \quad \textrm{where} 
    \label{eq:f_rsde}\\
    \tfb_\ph(\z_t, t, \x) &= f_\ph(\z_t, t, \x) - \frac{g_\ph^2(t)}{2} \n_{\z_t} \log q_\ph(\z_t|\x).\nonumber
\end{align}

Secondly, leveraging this reverse \gls{sde}~(see \cref{eq:f_rsde}) we incorporate a prediction of $\x$:
\begin{align}
    \label{eq:r_rsde} 
    d \z_t = \hf_{\th, \ph}(\z_t, t) d t + g_\ph(t) d \bw, 
    \quad \textrm{where} \quad
    \hf_{\th, \ph}(\z_t, t) = \tfb_\ph \big( \z_t, t, \hx_\th(\z_t, t) \big),
\end{align}
and $\hx_\th: \R^D \times [0,1] \mapsto \R^D$ is a function that predicts the data point $\x$. This \gls{sde} defines the dynamics of the generative trajectories $\{\z(t)\}_{t \in [0, 1]}$.

To fully specify the \gls{rproc}, it is also necessary to define a prior distribution $p(\z_1)$ and a reconstruction distribution $p(\x|\z_0)$. In all of our experiments, we set the prior $p(\z_1)$ to be a unit Gaussian distribution $\N(\z_1;0,I)$ and let $p(\x|\z_0)$ be a Gaussian distribution with a small variance $\N(\x;\z_0,\d^2 I)$, where $\d^2=10^{-4}$.

The above parameterization of the \gls{rproc} is not the only possibility. However, it is a convenient choice as it allows for the definition of the \gls{rproc} simply through the prediction of $\x$, similar to conventional diffusion models \cite{ho2020denoising}. We leave exploration of alternate parameterizations for future work.

\subsection{Optimization and Sampling}
\label{sec:nfdm_optimization}

With $F_\ph$ in \cref{eq:f} parameterized such that $q_\ph(\z_0|\x) \approx \d (\x - \z_0)$ and $q_\ph(\z_1|\x) \approx p(\z_1)$, we propose to optimize the forward and reverse processes of \gls{nfdm} jointly, minimizing the following objective:
\begin{align}
    \L = \E_{u(t)q(\x)q_\ph(\z_t \mid \x)} \left[ \frac{1}{2 g_\ph^2(t)} \big\| \tfb_\ph(\z_t, t, \x) - \hf_{\th, \ph}(\z_t, t) \big\|_2^2 \right].
    \label{eq:objective}
\end{align}

Since the \gls{fproc} is parameterized by $\ph$, we need to optimize the objective~(\cref{eq:objective}) with respect to both $\ph$ and $\th$ jointly. We discuss parameterization of the \gls{fproc} in \Cref{app:nfdm_parameterization}.

As we demonstrate in \Cref{app:derivation_nfdm_objective} the objective $\L$ shares similarities with the standard diffusion model objective \cite{ho2020denoising}, and it provides a variational bound on the model's log-likelihood $\log p_{\th, \ph}(\x)$. The objective $\L$ also exhibits strong connections with both the Flow Matching \cite{lipman2023flow} and Score Matching \cite{vincent2011connection} objectives. We explore these connections in \Cref{app:connections_objective}, where we also discuss the role of $g_\ph$ in detail. It is important to note that despite the parameterization of the \gls{rproc} through the prediction of $\x$, the objective $\L$ optimizes the generative dynamics and does not necessarily lead to accurate predictions of $\x$.

A key characteristic of the \gls{nfdm} objective is its compatibility with the \gls{simfree} paradigm, which is critical for efficient optimization. We summarize the training procedure in \Cref{alg:optimization}.

\begin{figure}[!t]
\begin{minipage}{0.49\textwidth}
\begin{algorithm}[H]
\caption{Optimization of \gls{nfdm}}
\label{alg:optimization}
\begin{algorithmic}
    \Require $q(\x)$, $F_\ph$, $g_\ph$, $\hx_\th$
    \For{learning iterations}
        \State $\x \sim q(\x)$, $t \sim u(t)$
        \State $\z_t \sim q_\ph(\z_t|\x)$
        \State $\L = \frac{1}{2 g_\ph^2(t)} \big\| \tfb_\ph(\z_t, t, \x) - \hf_{\th, \ph}(\z_t, t) \big\|_2^2$
        \State Gradient step on $\th$ and $\ph$ w.r.t. $\L$
    \EndFor
\end{algorithmic}
\end{algorithm}
\end{minipage}
\hfill
\begin{minipage}{0.49\textwidth}
\begin{algorithm}[H]
\caption{Stochastic Sampling from \gls{nfdm}}
\label{alg:sampling}
\begin{algorithmic}
    \Require $F_\ph$, $g_\ph$, $\hx_\th$, $T$ -- number of steps
    \State $\D t = \frac{1}{T}$, $\z_1 \sim p(\z_1)$
    \For{$t = 1, \dots, \frac{2}{T}, \frac{1}{T}$}
        \State $\bw \sim \N(0, I)$
        \State $\z_{t - \D t} = \z_t - \hf_{\th, \ph}(\z_t, t) \D t + g_\ph(t) \bw \sqrt{\D t}$
    \EndFor
    \State $\x \sim p(\x|\z_0)$
\end{algorithmic}
\end{algorithm}
\end{minipage}
\end{figure}

To sample from the trained \gls{rproc} we can simulate the \gls{sde}, as defined in \Cref{sec:nfdm_reverse}. This procedure is summarized in \Cref{alg:sampling}. Additionally, during the sampling process, we can adjust the level of stochasticity by modifying $g_\ph(t)$~(\cref{eq:f_sde}). It is important to note that changes to $g_\ph(t)$ also influence $\tff_\ph$~(\cref{eq:f_sde}) and $\hf_{\th, \ph}$~(\cref{eq:r_rsde}). In the extreme case where $g_\ph(t) \equiv 0$, the \gls{rproc} becomes deterministic, allowing us to utilize off-the-shelf numerical \gls{ode} solvers and to estimate densities \cite{chen2018neural, grathwohl2018scalable}. We provide an extended sampling discussion in \Cref{app:nfdm_sampling}.

\section{Neural Flow Bridge Models}
\label{sec:nfbm}

In this section we discuss a simple modification to the \gls{nfdm} framework that enables us to learn bridges between two data distributions, $q(\x_0)$ and $q(\x_1)$, a modification we refer to as \gls{nfbm}.

In the context of bridges, we consider a joint data distribution $q(\x_0, \x_1)$ (which may be factorized as $q(\x_0) q(\x_1)$ if paired data is unavailable). Our goal is to learn a generative process that starts from $\x_1$ and generates $\x_0$ such that $q(\x_0) \approx p_\th(\x_0) = \int q(\x_1) p_\th(\x_0|\x_1) d \x_1$. To turn \gls{nfdm} into \gls{nfbm}, we modify both the forward and reverse processes. In \gls{nfdm}, the \gls{fproc} is defined by two functions: $F_\ph(\e, t, \x)$~(see \cref{eq:f}) and $g_t$~(see \cref{eq:f_sde}). For \gls{nfbm}, we let $F_\ph$ depend on both $\x_0$ and $\x_1$, thereby conditioning the entire \gls{fproc} on these data points:
\begin{align}
    \z_t = F_\ph(\e, t, \x_0, \x_1),
    \label{eq:nfbm_f}
\end{align}

Similar to the discussion in \Cref{sec:nfdm_reverse}, for the \gls{rproc} of \gls{nfbm}, we predict the data points $\x_0$ and $\x_1$ from $\z_t$ and $t$, and substitute them into the conditional reverse time \gls{sde}~(\cref{eq:f_rsde}):
\begin{align}
    \label{eq:nfbm_r_rsde} 
    d \z_t = \hf_{\th, \ph}(\z_t, t) d t + g_\ph(t) d \bw, 
    \quad \textrm{where} \quad
    \hf_{\th, \ph}(\z_t, t) = \tfb_\ph \big( \z_t, t, \hx^{0,1}_\th(\z_t, t) \big).
\end{align}
Here, the function $\hx^{0,1}_\th(\z_t, t)$ returns predictions of both $\x_0$ and $\x_1$. Alternatively, the data point $\x_1$ can be reused as conditioning for intermediate steps of the \gls{rproc} instead of predicting both points, a strategy further detailed in \Cref{app:nfbm_reverse}.

When $F_\ph$ in \cref{eq:nfbm_f} is parameterized such that $q_\ph(\z_0|\x_0,\x_1) \approx \d (\x - \z_0)$ and $q_\ph(\z_1|\x_0,\x_1) \approx p(\z_1|\x_1)$ (see \Cref{app:nfbm_forward} for parameterization details of the \gls{nfbm} \gls{fproc}), we propose training \gls{nfbm} by minimizing the following objective:
\begin{align}
    \L = \E_{u(t)q(\x_0, \x_1)q_\ph(\z_t \mid \x_0, \x_1)} \left[ \frac{1}{2 g_\ph^2(t)} \big\| \tfb_\ph(\z_t, t, \x_0, \x_1) - \hf_{\th, \ph}(\z_t, t) \big\|_2^2 \right].
    \label{eq:nfbm_objective}
\end{align}

The derivation of this objective is provided in \Cref{app:derivation_nfbm_objective}. This objective shares key properties with the \gls{nfdm} objective, it provides a variational bound on the model's log-likelihood $\log p_{\th,\ph}(\x_0)$, and is compatible with the \gls{simfree} paradigm. Furthermore, it allows for sampling with various levels of stochasticity (by adjusting $g_\ph(t)$) including deterministic sampling (when $g_\ph(t) \equiv 0$).

Thus, the \gls{nfdm} framework not only enables the construction of generative models, but also facilitates learning bridge models between two data distributions.

\section{Restricted NFDM}
\label{sec:restricted}

We introduce \gls{nfdm} as a powerful framework that enables learning of the \gls{fproc}. However, there is in general an infinite number of forward and reverse processes that correspond to each other. In this section we discuss how the flexibility of \gls{nfdm} (and \gls{nfbm}) allows learning generative dynamic with user-specified beneficial properties.

Suppose our objective is to learn straight generative \gls{ode} trajectories, which can be highly beneficial, as it enables generation with far fewer steps in the \gls{ode} solver. One approach is to introduce penalties on the curvature of the \gls{rproc}'s trajectories and let the learnable \gls{fproc} to adapt to align with the \gls{rproc}. As a result, we would obtain a generative process that not only corresponds to the \gls{fproc}, ensuring accurate data  generation, but also features the desired property of straight trajectories. We discuss \gls{nfdm} with restrictions in more details in \Cref{app:restrictions_discussion}.

We propose learning a model with curvature penalty as suggested by \cite{kelly2020learning}:
\begin{align}
    \Lot = \L + \l \Lcrv, 
    \quad \textrm{where} \quad
    \Lcrv = \E_{u(t) q_\ph(\x, \z_t)} \left\| \frac{d \hf_{\th, \ph}(\z_t, t)}{d t} \right\|_2^2.
    \label{eq:l_crv}
\end{align}

We refer to this variant as \gls{nfdm}-OT\footnote{OT stands for optimal transport. This designation is used for convenience. While straight trajectories are a necessary condition for dynamic optimal transport, they are not sufficient.}. $\Lcrv$ is an additional curvature loss that penalizes the second time derivative of the generative \gls{ode} trajectories. $\L$ is estimated as in \Cref{sec:nfdm_optimization}, whereas when calculating $\Lcrv$ we set $g_\ph(t) \equiv 0$. $\Lcrv \equiv 0$ ensures that the generative trajectories are straight lines. In our experiments, we set $\l = 10^{-2}$. Empirical evidence suggests that higher values of $\l$ lead to slower convergence of the model. The specifics of the curvature loss are elaborated upon in more detail in \Cref{app:restrictions_ot}.

We would like to emphasise that the purpose of this section is to provide an example of how the \gls{nfdm} framework allows learning dynamics with specific properties. Not to propose the best way for learning generative straight-line generative dynamics. In addition, we note that conventional diffusion models are incapable of handling such penalization strategies. In diffusion models with a fixed \gls{fproc}, the target for the \gls{rproc} is predetermined and the corresponding \gls{ode} trajectories are highly curved. Hence, imposing constraints on the \gls{rproc}, such as trajectory straightness, would lead to a mismatch with the \gls{fproc} and, consequently, an inability to generate samples with high data fidelity.

\section{Experiments}
\label{sec:experiments}

We first showcase results demonstrating that \gls{nfdm} consistently achieves better likelihood compared to baselines, obtaining state-of-the-art diffusion modeling results on the CIFAR-10 \cite{krizhevsky2009learning} and downsampled ImageNet \cite{deng2009imagenet, van2016pixel} datasets. Then, we explore the \gls{nfdm}-OT modification, which penalizes the curvature of the deterministic generative trajectories. The \gls{nfdm}-OT reduces trajectory curvature, significantly lowering the number of generative steps required for sampling. Finally, we demonstrate the \gls{nfbm} modification, learning bridges on the AFHQ \cite{choi2020stargan} dataset and several synthetic examples.

We report the \gls{nll} in \gls{bpd} and sample quality measured by the \gls{fid} score \cite{heusel2017gans}. The \gls{nll} is calculated by integrating the \glspl{ode} using the RK45 solver \cite{dormand1980family}, with all \gls{nll} metrics computed on test data. For \gls{fid}, we provide the average over 50k generated images.

Unless otherwise stated, we parameterized the \gls{fproc} of \gls{nfdm} and \gls{nfbm} such that $q_\ph(\z_t|\x)$ is a Gaussian with learnable mean and covariance (see details in \Cref{app:nfdm_parameterization,app:nfbm_forward}). For a detailed description of parameterizations and other experimental details, please refer to \Cref{app:implementation}.

The primary aim of the experiments with \gls{nfdm}-OT and \gls{nfbm} is not to introduce a novel model that surpasses others in few-step generation or learning bridges, but rather to showcase the ability of the \gls{nfdm} framework to learn generative dynamics with specific properties. The straightness of the trajectories is just one example of such a property. We leave it for future research to explore different parameterizations and  modified objectives for \gls{nfdm} that may yield even better results.

Please refer to \cref{app:additional_samples} for generated samples. The code is available at \url{https://github.com/GrigoryBartosh/neural_diffusion}.

\subsection{Likelihood Estimation}
\label{sec:experiments_nll}

\begin{table*}[!t]
\caption{Comparison of \gls{nfdm} results with baselines on density estimation tasks. We present results in terms of \acrshort{bpd}, lower is better. \gls{nfdm} achieves state-of-the-art results across all three benchmark tasks.}
\label{tab:nll}
\centering
\begin{tabular}{llccc}
\toprule
Model &  & \textbf{CIFAR10} & \textbf{ImageNet 32} & \textbf{ImageNet 64}  \\ \cmidrule(r){1-1} \cmidrule(r){3-5}
\cellcolor[HTML]{EFEFEF}DDPM \cite{ho2020denoising} & \cellcolor[HTML]{EFEFEF} & \cellcolor[HTML]{EFEFEF}$3.69$ & \cellcolor[HTML]{EFEFEF} & \cellcolor[HTML]{EFEFEF}\\
Score SDE \cite{song2021scorebased} &  & $2.99$ &  & \\
\cellcolor[HTML]{EFEFEF}Improved DDPM \cite{nichol2021improved} & \cellcolor[HTML]{EFEFEF} & \cellcolor[HTML]{EFEFEF}$2.94$ & \cellcolor[HTML]{EFEFEF} & \cellcolor[HTML]{EFEFEF}$3.54$ \\
VDM \cite{kingma2021variational} &  & $2.65$ & $3.72$ & $3.40$ \\
\cellcolor[HTML]{EFEFEF}Score Flow \cite{song2021maximum} & \cellcolor[HTML]{EFEFEF} & \cellcolor[HTML]{EFEFEF}$2.83$ & \cellcolor[HTML]{EFEFEF}$3.76$ & \cellcolor[HTML]{EFEFEF} \\
Flow Matching \cite{lipman2023flow} &  & $2.99$ & $3.53$ & $3.31$ \\
\cellcolor[HTML]{EFEFEF}Stochastic Interp. \cite{albergo2023building} & \cellcolor[HTML]{EFEFEF} & \cellcolor[HTML]{EFEFEF}$2.99$ & \cellcolor[HTML]{EFEFEF}$3.48$ & \cellcolor[HTML]{EFEFEF} \\
i-DODE \cite{zheng2023improved} &  & $2.56$ & $3.43$ & \\
\cellcolor[HTML]{EFEFEF}NDM \cite{bartosh2023neural} & \cellcolor[HTML]{EFEFEF} & \cellcolor[HTML]{EFEFEF}$2.70$ & \cellcolor[HTML]{EFEFEF}$3.55$ &  \cellcolor[HTML]{EFEFEF} $3.35$ \\
MuLAN \cite{sahoo2024diffusion} &  & $2.55$ & $3.67$ & \\
\cellcolor[HTML]{EFEFEF}\gls{nfdm} (\textbf{Gaussian} $q_\ph(\z_t|\x)$) & \cellcolor[HTML]{EFEFEF} & \cellcolor[HTML]{EFEFEF}$2.49$ & \cellcolor[HTML]{EFEFEF}$3.36$ & \cellcolor[HTML]{EFEFEF}$\mathbf{3.20}$ \\
\gls{nfdm} (\textbf{non-Gaussian} $q_\ph(\z_t|\x)$) &  & $\mathbf{2.48}$ & $\mathbf{3.34}$ & $\mathbf{3.20}$\\
\bottomrule
\end{tabular}
\end{table*}

\begin{figure}[tp]
    \centering
    \parbox[b]{.49\textwidth}{
        \begin{subfigure}[b]{\linewidth}
            \includegraphics[width=\textwidth]{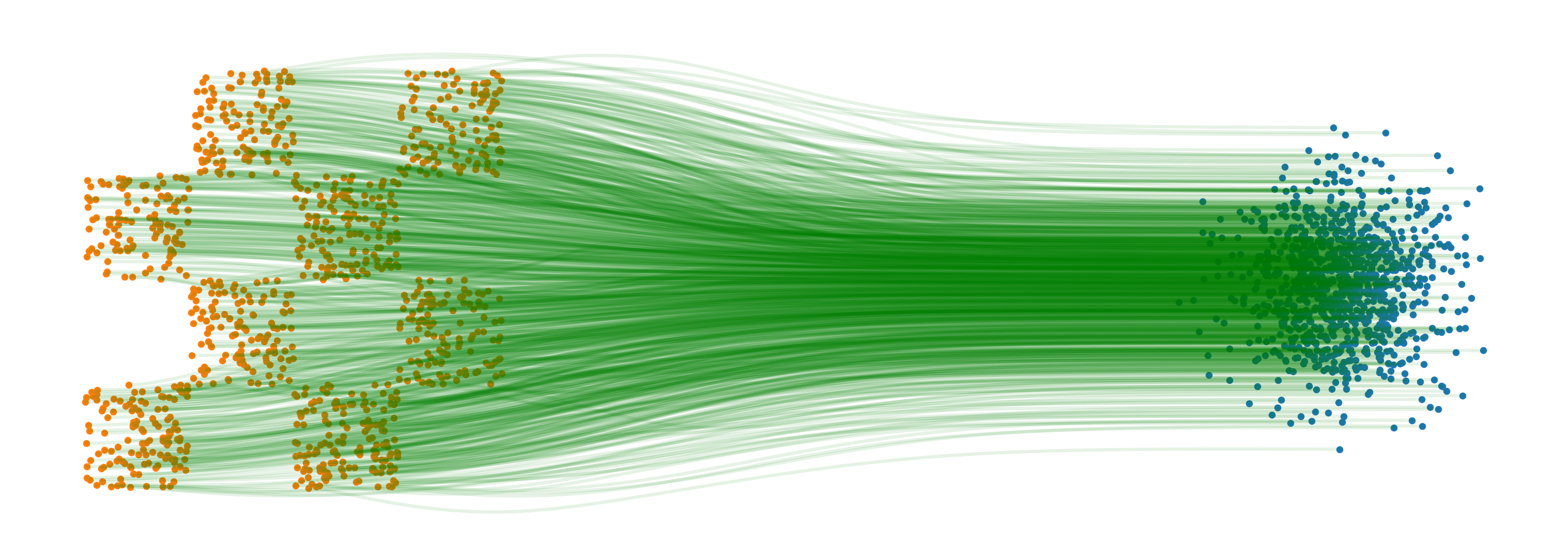}
            \caption{Score SDE \cite{song2021scorebased}}
            \label{fig:traj_ddpm}
        \end{subfigure}
    }
    \hfill
    \parbox[b]{.49\textwidth}{
        \begin{subfigure}[b]{\linewidth}
            \includegraphics[width=\textwidth]{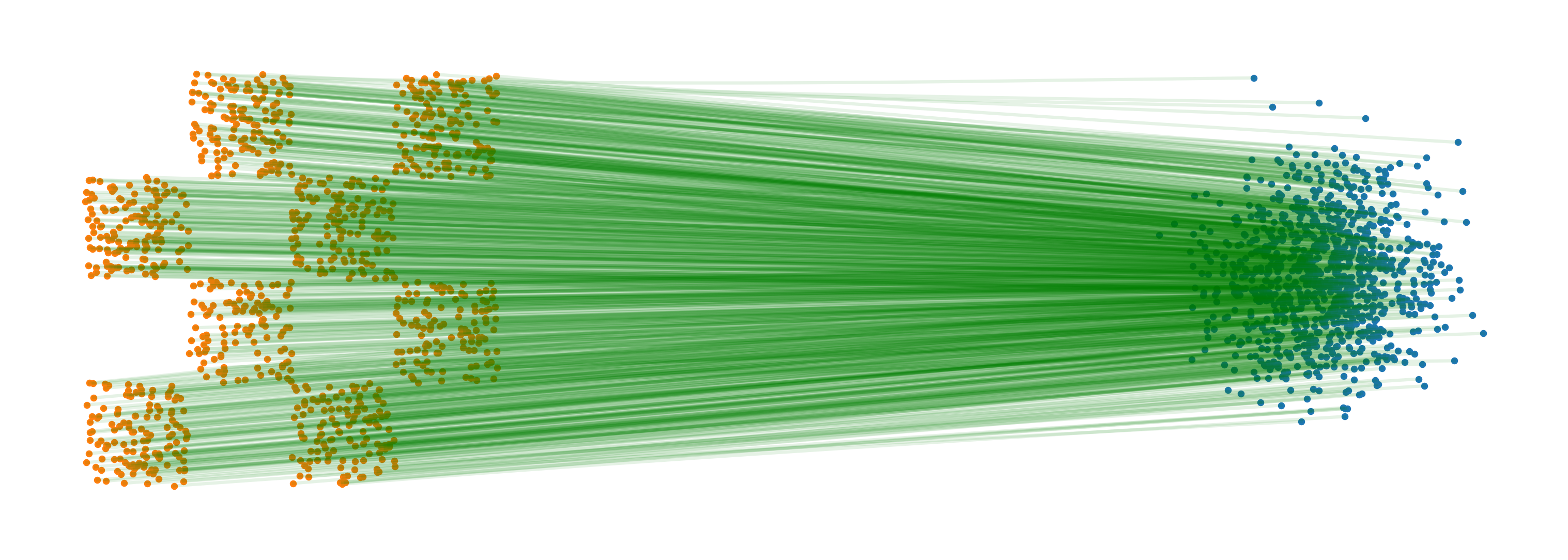}
        \caption{\gls{nfdm}-OT (this paper)}
        \label{fig:traj_nfdm_ot}
        \end{subfigure}
    }
    \caption{Comparison of trajectories between the data distribution (on the left) and the prior distribution (on the right), as learned by conventional diffusion and \gls{nfdm}-OT.}
\end{figure}

For the first experiments, in addition to Gaussian we also provide results with non-Gaussian parameterization of the \gls{fproc} (see \Cref{app:nfdm_parameterization}). \Cref{tab:nll} summarizes the \gls{nll} results on the CIFAR-10 and two downsampled ImageNet datasets. Notably, \gls{nfdm} outperforms diffusion-based baselines on all three datasets with both parameterizations, achieving state-of-the-art performance. The improved performance due to \gls{nfdm} is a natural progression for the following reasons.

First, it is well-established \cite{song2021maximum, zheng2023improved} that diffusion models exhibit improved likelihood estimation when trained with the full \gls{elbo} objective. The objective $\L$~(\cref{eq:objective}) used for training the \gls{nfdm} is also a variational bound on the likelihood.

Second, diffusion models can be seen as hierarchical \glspl{vae}. From this perspective, most baselines in \Cref{tab:nll} resemble \glspl{vae} with either fixed or constrained variational distributions. In contrast, the \gls{nfdm} extends beyond these baselines by providing a more flexible variational distribution. That is true even for the Gaussian parameterization with a  mean and covariance that is non-linear in $\x$ and $t$. This flexibility allows the \gls{nfdm} to better conform to the \gls{rproc}, consequently enhancing likelihood estimation.

\subsection{Straight Trajectories}
\label{sec:experiments_ot}

We next evaluate \gls{nfdm}-OT, designed to penalize the deterministic generative trajectory curvature. First, we compare \gls{nfdm}-OT with a conventional continuous-time diffusion model. \Cref{fig:traj_ddpm} illustrates deterministic trajectories between a two-dimensional data distribution and a unit Gaussian distribution learnt by a conventional diffusion model \cite{song2021scorebased}. \Cref{fig:traj_nfdm_ot} depicts trajectories learnt by \gls{nfdm}-OT. Conventional diffusion, being constrained in its \gls{fproc}, learns highly curved trajectories, whereas \gls{nfdm}-OT successfully learns straight generative trajectories as desired. In \Cref{app:additional_nfdm_ot_forward} we provide some additional reult, demonstrating the importance of learnable \gls{fproc} for \gls{nfdm}-OT.

Then, we present results of \gls{nfdm}-OT on image datasets. \Cref{tab:ot} reports the \gls{fid} scores for $2$, $4$, and $12$ \gls{nfe} with respect to the function $\hf_{\th, \ph}$~(\cref{eq:r_rsde}). In this experiment, we employ Euler's method for sampling integration. For the specified \glspl{nfe}, \gls{nfdm}-OT demonstrates superior sample quality compared to other approaches with similar \gls{nfe} values. Specifically, \gls{nfdm}-OT outperforms approaches specifically designed to minimize the curvature of generative trajectories \cite{pooladian2023multisample,lee2023minimizing}.

Importantly, \gls{nfdm}-OT is trained with an \gls{elbo}-based objective~(\cref{eq:objective}), which is known to yield higher \gls{fid} scores for diffusion models \cite{song2021maximum, zheng2023improved}. In contrast, some of the approaches listed in \Cref{tab:ot} are trained with different objectives, leading to improved \gls{fid} scores. Even so, for comparable \gls{nfe} values \gls{nfdm}-OT still achieves superior results. We provide additional results of \gls{nfdm}-OT in \Cref{app:additional_nfdm_images}.

\begin{table*}[!t]
\caption{Summary of \gls{fid} results for few-step generation. The table is divided into three sections, based on different types of methods: those that do not minimize curvature, solvers for pretrained models, and models that specifically aim to minimize curvature. For the DDPM, we include results corresponding to two distinct objectives: the full \gls{elbo}-based objective and a simplified objective ($\Lsimple$). \gls{nfdm}-OT outperforms baselines with comparable \acrshort{nfe} values.}
\label{tab:ot}
\centering
\small
\begin{tabular}{llcccccccc}
\toprule
 &  & \multicolumn{2}{c}{\textbf{CIFAR-10}} &  & \multicolumn{2}{c}{\textbf{ImageNet 32}} &  &  \multicolumn{2}{c}{\textbf{ImageNet 64}}  \\
Model &  & \acrshort{nfe} $\downarrow$ & \acrshort{fid} $\downarrow$ &  & \acrshort{nfe} $\downarrow$ & \acrshort{fid} $\downarrow$ &  & \acrshort{nfe} $\downarrow$ & \acrshort{fid} $\downarrow$ \\ \cmidrule(r){1-1} \cmidrule(r){3-4} \cmidrule(r){6-7} \cmidrule(r){9-10}
DDPM ($\Lsimple$) \cite{ho2020denoising} &  & $1000$ & $3.17$ &  &  &  &  &  &  \\
\cellcolor[HTML]{EFEFEF}DDPM (\gls{elbo}) \cite{ho2020denoising} & \cellcolor[HTML]{EFEFEF} & \cellcolor[HTML]{EFEFEF}$1000$ & \cellcolor[HTML]{EFEFEF}$13.51$ & \cellcolor[HTML]{EFEFEF} & \cellcolor[HTML]{EFEFEF} & \cellcolor[HTML]{EFEFEF} & \cellcolor[HTML]{EFEFEF} & \cellcolor[HTML]{EFEFEF} & \cellcolor[HTML]{EFEFEF} \\
Flow Matching \cite{lipman2023flow} &  & $142$ & $6.35$ &  & $122$ & $5.02$ &  & $138$ & $14.14$ \\ \midrule
\cellcolor[HTML]{EFEFEF}DDIM \cite{song2021denoising} & \cellcolor[HTML]{EFEFEF} & \cellcolor[HTML]{EFEFEF}$10$ & \cellcolor[HTML]{EFEFEF}$13.36$ & \cellcolor[HTML]{EFEFEF} & \cellcolor[HTML]{EFEFEF} & \cellcolor[HTML]{EFEFEF} & \cellcolor[HTML]{EFEFEF} & \cellcolor[HTML]{EFEFEF} & \cellcolor[HTML]{EFEFEF} \\
 &  & $12$ & $5.28$ &  &  &  &  &  &  \\
\multirow{-2}{*}{DPM Solver \cite{lu2022dpm}} &  & $24$ & $2.75$ &  &  &  &  &  &  \\ \midrule
\cellcolor[HTML]{EFEFEF}Trajectory Curvature Minimization \cite{lee2023minimizing} & \cellcolor[HTML]{EFEFEF} & \cellcolor[HTML]{EFEFEF}$5$ & \cellcolor[HTML]{EFEFEF}$18.74$ & \cellcolor[HTML]{EFEFEF} & \cellcolor[HTML]{EFEFEF} & \cellcolor[HTML]{EFEFEF} & \cellcolor[HTML]{EFEFEF} & \cellcolor[HTML]{EFEFEF} & \cellcolor[HTML]{EFEFEF} \\
 &  &  &  &  & $4$ & $17.28$ &  & $4$ & $38.45$ \\
\multirow{-2}{*}{Multisample Flow Matching \cite{pooladian2023multisample}} &  &  &  &  & $12$ & $7.18$ &  & $12$ & $17.6$ \\
\cellcolor[HTML]{EFEFEF} & \cellcolor[HTML]{EFEFEF} & \cellcolor[HTML]{EFEFEF}$2$ & \cellcolor[HTML]{EFEFEF}$12.44$ & \cellcolor[HTML]{EFEFEF} & \cellcolor[HTML]{EFEFEF}$2$ & \cellcolor[HTML]{EFEFEF}$9.83$ & \cellcolor[HTML]{EFEFEF} & \cellcolor[HTML]{EFEFEF}$2$ & \cellcolor[HTML]{EFEFEF}$27.70$ \\
\cellcolor[HTML]{EFEFEF} & \cellcolor[HTML]{EFEFEF} & \cellcolor[HTML]{EFEFEF}$4$ & \cellcolor[HTML]{EFEFEF}$7.76$ & \cellcolor[HTML]{EFEFEF} & \cellcolor[HTML]{EFEFEF}$4$ & \cellcolor[HTML]{EFEFEF}$6.13$ & \cellcolor[HTML]{EFEFEF} & \cellcolor[HTML]{EFEFEF}$4$ & \cellcolor[HTML]{EFEFEF}$17.28$ \\
\cellcolor[HTML]{EFEFEF}\multirow{-3}{*}{\gls{nfdm}-OT (\textbf{this paper})} & \cellcolor[HTML]{EFEFEF} & \cellcolor[HTML]{EFEFEF}$12$ & \cellcolor[HTML]{EFEFEF}$5.20$ & \cellcolor[HTML]{EFEFEF} & \cellcolor[HTML]{EFEFEF}$12$ & \cellcolor[HTML]{EFEFEF}$4.11$ & \cellcolor[HTML]{EFEFEF} & \cellcolor[HTML]{EFEFEF}$12$ & \cellcolor[HTML]{EFEFEF}$11.58$ \\
\bottomrule
\end{tabular}
\end{table*}

\begin{figure}[tp]
    \centering
    \parbox[b]{.445\textwidth}{
        \begin{subfigure}[b]{\linewidth}
            \includegraphics[width=\textwidth]{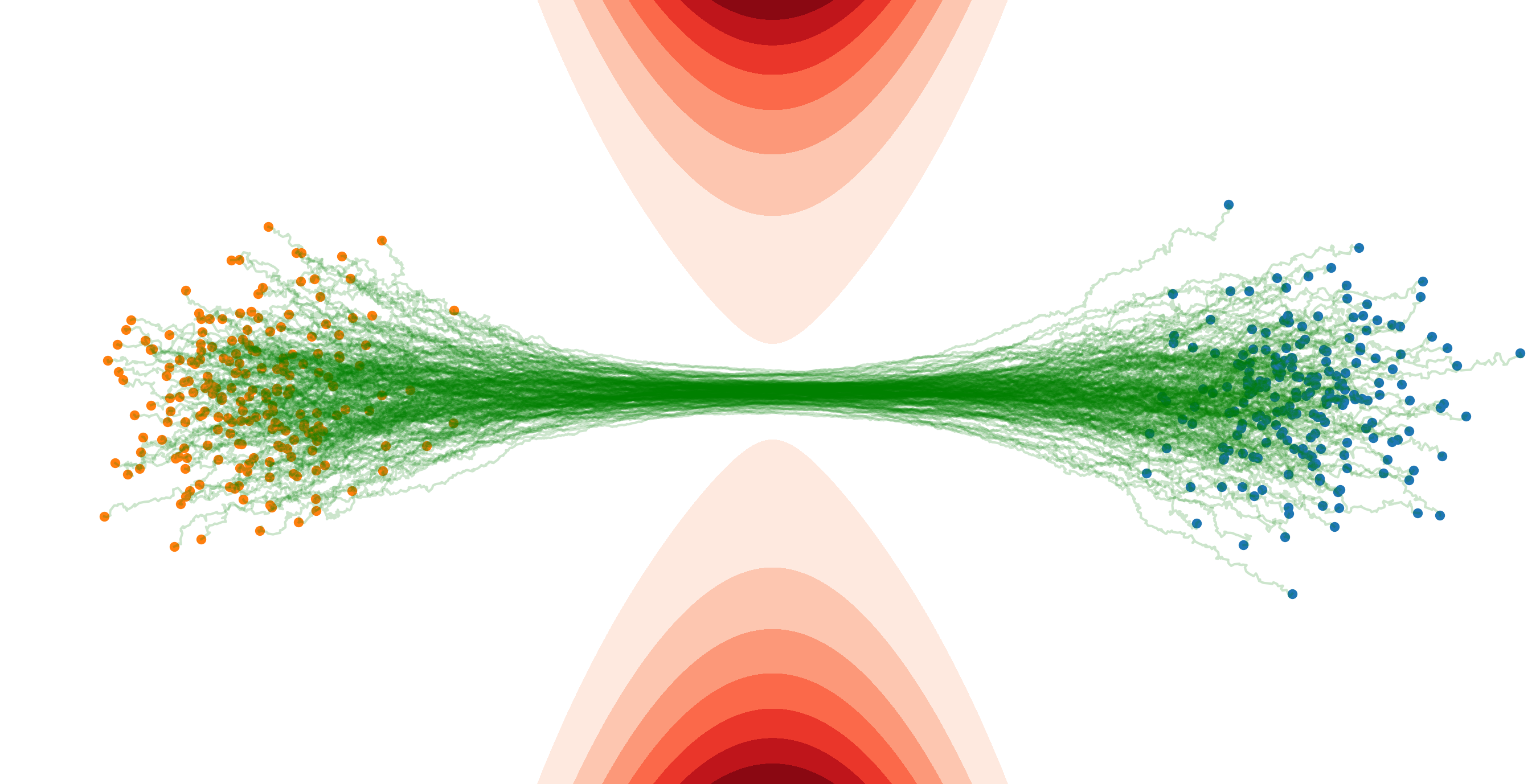}
        \end{subfigure}
    }
    \hfill
    \parbox[b]{.548\textwidth}{
        \begin{subfigure}[b]{\linewidth}
            \includegraphics[width=\textwidth]{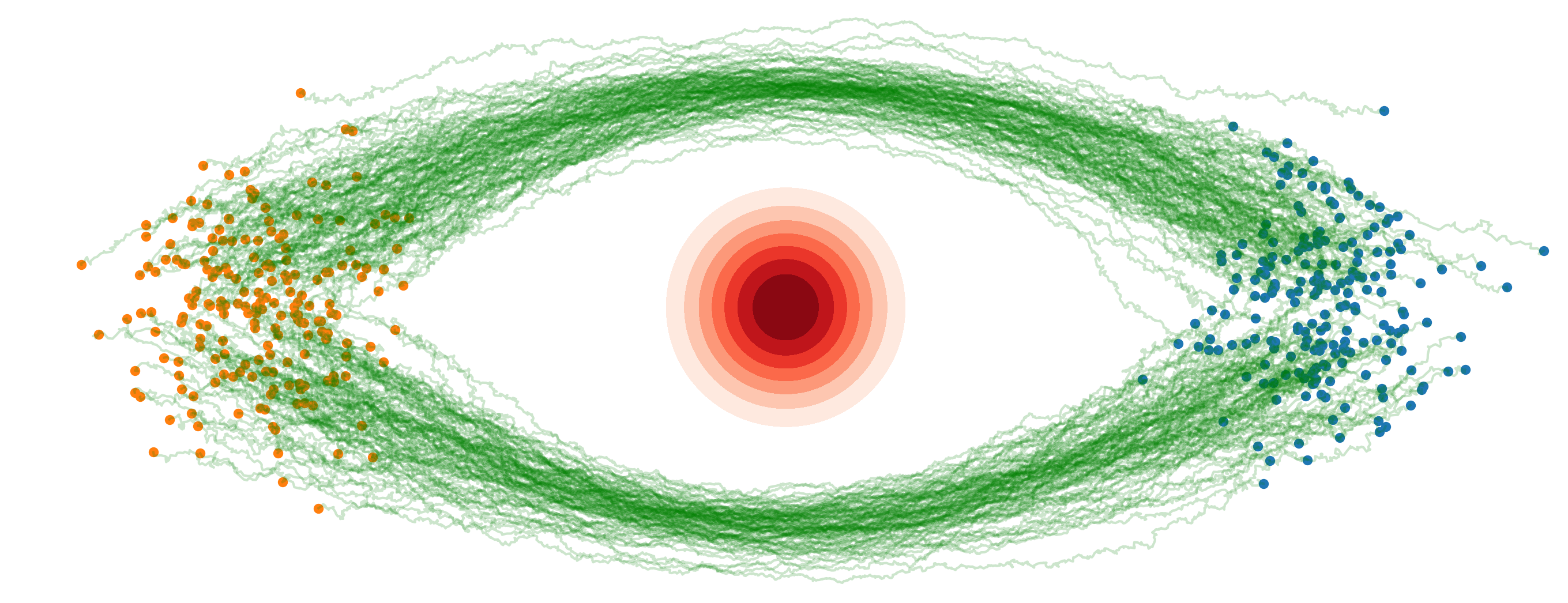}
        \end{subfigure}
    }
    \caption{Trajectories of \gls{nfbm} between two-dimensional data distributions, trained with an additional penalty to avoid obstacles.}
    \label{fig:nfbm}
\end{figure}

\newpage
\subsection{Bridges}
\label{sec:experiments_bridges}

\begin{wraptable}[9]{r}{0.3\textwidth}
\centering
\captionof{table}{\acrshort{fid} values of dog $\rightarrow$ cat on AFHQ $64$.}
\label{tab:bridge}
\centering
\begin{tabular}{llc}
\toprule
Model &  & \acrshort{fid} $\downarrow$  \\ \cmidrule(r){1-1} \cmidrule(r){3-3}
\cellcolor[HTML]{EFEFEF}DSBM \cite{shi2024diffusion} & \cellcolor[HTML]{EFEFEF} & \cellcolor[HTML]{EFEFEF}$14.16$ \\
GSBM \cite{liu2024generalized} &  & $12.39$ \\
\cellcolor[HTML]{EFEFEF}\acrshort{nfbm} & \cellcolor[HTML]{EFEFEF} & \cellcolor[HTML]{EFEFEF} $\mathbf{12.06}$ \\
\bottomrule
\end{tabular}
\end{wraptable}

We consider the task of translating images of dogs into images of cats from the downsampled AFHQ dataset. \Cref{tab:bridge} reports the \gls{fid} scores on images sampled with $1000$ steps. Notably, \gls{nfbm} outperforms the baselines, demonstrating the effectiveness of our framework for learning bridges.

Finally, we demonstrate that the framework flexibility also allows for learning dynamics with specific properties in the case of \gls{nfbm}. To illustrate this, we adopted the strategy discussed in \Cref{sec:restricted} and learn \gls{nfbm} with an additional penalty to avoid obstacles (see details in \Cref{app:implementation_obstacles}). \Cref{fig:nfbm} visualizes the learned stochastic trajectories. It is evident that \gls{nfbm} has efficiently learned to translate distributions while avoiding obstacles.

\section{Related Work}
\label{sec:related_work}

Diffusion models, originally proposed by \cite{sohl2015deep}, have evolved significantly through subsequent developments \cite{song2019generative, ho2020denoising}. These advancements have resulted in remarkable generative quality for high-dimensional data distributions \cite{dhariwal2021diffusion, saharia2022photorealistic}. Nevertheless, conventional diffusion models, typically relying on a linear Gaussian \gls{fproc}, may not optimally fit some data distributions. To address this, alternative \glspl{fproc} have been explored, such as alternative linear transformations \cite{du2023flexible, singhal2023where}, combining blurring with Gaussian noise injection \cite{rissanen2023generative, daras2022soft, hoogeboom2023blurring}, diffusion in the wavelet spectrum \cite{phung2023wavelet}, and forward processes based on the exponential family \cite{okhotin2024star}. These models are limited by their fixed \glspl{fproc} and may be seen as specific instances of \gls{nfdm}.

Some studies have focused on making the \gls{fproc} learnable. Approaches include a learnable noise injection schedule \cite{kingma2021variational, nichol2021improved, sahoo2024diffusion} and learning data transformations like time-independent transformations based on \glspl{vae} \cite{vahdat2021score, rombach2022high} and normalizing flows \cite{kim2022maximum}, or time-dependent transformations \cite{gu2023fdm, singhal2023where, nielsen2024diffenc, bartosh2023neural}. These methods can also be considered special cases of \gls{nfdm} with specific transformations $F_\ph$~(\cref{eq:f}).

To improve the sampling efficiency, in orthogonal line of works alternative sampling  methods have been studied \cite{tachibana2021taylor, liu2022pseudo, lu2022dpm, shaul2024bespoke}. Additionally, distillation techniques have been applied \cite{salimans2022progressive, song2023consistency, zheng2023fast, yin2023one} to enhance sampling speed, albeit at the cost of training additional models. Most of these approches are compatible with \gls{nfdm} and may be combined for further gains.

Another line of works proposed technics for learning straighter generative trajectories. However, these approaches requires distillation \cite{liu2023flow, liu2022rectified}, solving unstable min-max problem \cite{albergo2023stochastic}, estimating an optimal coupling over minibatches \cite{pooladian2023multisample, tong2023improving} of the entire dataset, which, for large datasets, may become uninformative, or can be viewed as specific instances \cite{lee2023minimizing, shaul2023kinetic} of \gls{nfdm}.

The connections between \gls{nfdm} and some related works is discussed further in \Cref{app:connections}.

\section{Conclusion and Limitations}
\label{sec:conclusion}

In this paper we introduced \gls{nfdm}, a novel \gls{simfree} framework for improved diffusion modeling through a learnable forward processes. \gls{nfdm} outperforms baseline diffusion models on standard benchmarks, showcasing its effectiveness. However, \gls{nfdm} is not just another model, but rather it is a versatile framework that facilitates the predefined specification and learning of a forward process. Similarly, the penalty on the curvature of deterministic trajectories discussed in \Cref{sec:restricted} represents just one specific example of the restrictions that \gls{nfdm} can accommodate.

Nonetheless, the advantages of \gls{nfdm} come with certain trade-offs.  Once the \gls{fproc} is parameterized using a neural network, this leads to approximately 2.2 times longer optimization iteration of \gls{nfdm} compared to conventional diffusion models. Additionally, as discussed in \Cref{sec:nfdm_forward}, \gls{nfdm} is limited to \gls{fproc} parameterisations with functions that match specific conditions.

Despite these challenges, we are optimistic about the potential of \gls{nfdm}. \gls{nfdm} opens up significant prospects for exploring new generative dynamics. We believe that with alternative parameterizations, modifications of the objective, and the integration of orthogonal approaches, \gls{nfdm} has the potential to achieve even better results across a range of data modalities.

\bibliography{bibliography}
\bibliographystyle{abbrv}


\newpage
\appendix

\section{Derivations}
\label{app:derivations}

\subsection{Derivation of NFDM's Objective}
\label{app:derivation_nfdm_objective}

Our goal is to derive a variational bound on the \acrlong{nll} of \gls{nfdm}. For the convenience of the following derivations we will consider the \gls{fproc} as a conditional reverse \gls{sde}~(\cref{eq:f_rsde}) that starts from $\z_1 \sim q_\ph(\z_1|\x)$ and flows backwards in time:
\begin{align}
    d \z_t &= \tfb_\ph(\z_t, t, \x) d t + g_\ph(t) d \bw.
    \label{eq:app_f_rsde}
\end{align}

Similarly to a conditional \gls{sde}~(\cref{eq:f_sde}) flowing forwards in time, this \gls{sde} also corresponds to conditional marginal distribution $q_\ph(\z_t|\x)$, which is implicitly defined by the invertible transformation $\z_t = F_\ph(\e, t, \x)$~(\cref{eq:f}), where $\e \sim q(\e)$.

For the \gls{rproc} we use the definition from \Cref{sec:nfdm_reverse} as reverse \gls{sde}~(\cref{eq:r_rsde}) that starts from the prior distribution $p(\z_1)$, flows backwards in time, and finally samples $\x \sim p(\x|\z_0)$:
\begin{align}
    d \z_t = & \hf_{\th, \ph}(\z_t, t) d t + g_\ph(t) d \bw.
    \label{eq:app_r_rsde}
\end{align}

To derive the \gls{nfdm}'s variational bound on the \gls{nll} we discretize processes, derive an objective in discrete case and then consider the limit of discretization to switch back to continuous time.

We discretize the conditional reverse \gls{sde} in \cref{eq:app_f_rsde} and the reverse \gls{sde} in \cref{eq:app_r_rsde}, transitioning from continuous-time trajectories $\{\z(t)\}_{t \in [0, 1]}$ to discrete-time trajectories $\bz_0, \bz_\frac{1}{T}, \dots, \bz_1$, where $T$ represents the number of discrete steps of size $\D t = \frac{1}{T}$.

The discrete conditional reverse process starts from distribution $\bq_\ph(\bz_1|\x) = q_\ph(\bz_1|\x)$ and follows conditional reverse Markov chain:
\begin{align}
    \bq_\ph( \bz_{t - \D t} | \bz_t, \x) = \N \Big( \bz_{t - \D t}; \bz_t - \D t \tfb_\ph(\bz_t, t, \x), \D t g^2_\ph(t) I \Big).
    \label{eq:app_f_rsde_discr}
\end{align}

Similarly, the discrete generative process shares distributions $\bp(\bz_1) = p(\bz_1)$ and $\bp(\x|\bz_0) = p(\x|\bz_0)$ with the continuous one. However, instead of following the reverse \gls{sde}~(\cref{eq:app_r_rsde}) it follows the reverse Markov chain:
\begin{align}
    \bp_{\th, \ph}( \bz_{t - \D t} | \bz_t) = \N \Big( \bz_{t - \D t}; \bz_t - \D t \hf_{\th, \ph}(\bz_t, t), \D t g^2_\ph(t) I \Big).
    \label{eq:app_r_rsde_discr}
\end{align}

It's important to note that once we discretize the trajectories, $\bq_\ph(\bz_t|\x) \neq q_\ph(\bz_t|\x)$ and $\bp_{\th, \ph}(\bz_t) \neq p_{\th, \ph}(\bz_t)$ except for $t=1$. However, this discretization corresponds to an Euler-Maruyama method of integrating \glspl{sde}. Thus, we know that as $\D t \rightarrow 0$, the discretized trajectories converge to the continuous ones.

Now, considering these discrete processes and recognizing that they are Markovian conditional and unconditional processes moving backwards in time, we can leverage a key result from \cite{sohl2015deep, ho2020denoising}:
\begin{align}
    & \E_{q(\x)} \left[ - \log \bp_{\th, \ph}(\x) \right] \leq \bLrec + \bLdiff + \bLprior, \quad \textrm{where} 
    \label{eq:app_discr_elbo}\\
    & \bLrec= \E_{\bq_\ph(\x, \bz_0)} \left[ - \log p(\x | \bz_0) \right], \\
    & \bLdiff = \sum_{t=\frac{1}{T}}^T \E_{\bq_\ph(\x, \bz_t)} \left[ \KL \big( 
    \bq_\ph( \bz_{t - \D t} | \bz_t, \x) \|
    \bp_{\th, \ph}( \bz_{t - \D t} | \bz_t)
    \big) \right], \\
    & \bLprior = \E_{q(\x)} \left[ \KL \big( \bq_\ph(\bz_1|\x) \| \bp(\bz_1) \big) \right].
\end{align}

\Cref{eq:app_discr_elbo} establishes a variational upper bound on the \acrlong{nll} of discretized \gls{rproc}. Taking the limits of this upper bound, the reconstruction term $\Lrec = \bLrec$ and the prior term $\Lprior = \bLprior$ retain their forms. However, we can rewrite the diffusion term:
\begin{align}
    \L_\mathrm{diff}
    & = \lim_{\D t \rightarrow 0} \bLdiff \\
    & = \lim_{\D t \rightarrow 0}
    \sum_{t=\frac{1}{T}}^T \E_{\bq_\ph(\x, \bz_t)} \left[ \KL \big( 
    \bq_\ph( \bz_{t - \D t} | \bz_t, \x) \|
    \bp_{\th, \ph}( \bz_{t - \D t} | \bz_t)
    \big) \right] \\
    & = \lim_{\D t \rightarrow 0}
    \sum_{t=\frac{1}{T}}^T \E_{\bq_\ph(\x, \bz_t)} \left[
    \frac{1}{2 \D t g^2_\ph(t)} \left\|
    \cancel{\bz_t} - \D t \hf_{\th, \ph}(\bz_t, t) - \cancel{\bz_t} + \D t \tfb_\ph(\bz_t, t, \x)
    \right\|_2^2
    \right] \\
    & = \lim_{\D t \rightarrow 0}
    \sum_{t=\frac{1}{T}}^T \E_{\bq_\ph(\x, \bz_t)} \left[
    \frac{\D t^{\cancel{2}}}{2 \cancel{\D t} g^2_\ph(t)} \left\|
    \tfb_\ph(\bz_t, t, \x) - \hf_{\th, \ph}(\bz_t, t)
    \right\|_2^2
    \right] \\
    & = \E_{u(t) \bq_\ph(\x, \bz_t)} \left[
    \frac{1}{2 g^2_\ph(t)} \left\|
    \tfb_\ph(\bz_t, t, \x) - \hf_{\th, \ph}(\bz_t, t)
    \right\|_2^2
    \right],
\end{align}
where $u(t)$ is a uniform distribution over a unit interval.

Therefore, as the discretized trajectory tends to the continuous one when $\D t \rightarrow 0$, we can use the obtained limits as a variational bound on the \acrlong{nll} of the continuous model:
\begin{align}
    \E_{q(\x)} \left[ - \log p_{\th, \ph}(\x) \right] \leq \L = \Lrec + \Ldiff + \Lprior.
    \label{app:elbo}
\end{align}

In general case, when function $F_\ph$ is actually parameterized by $\ph$, the objective $\L$ implies that we must optimise all three terms with respect to parameters $\ph$ and $\th$. However, when $F_\ph$ is parameterized such that $q_\ph(\z_0|\x) \approx \d (\x - \z_0)$ and $q_\ph(\z_1|\x) \approx p(\z_1)$ for any $\ph$, the reconstruction $\Lrec$ and prior $\Lprior$ terms are constants with respect to the parameters.

\subsection{Derivation of NFBM's Objective}
\label{app:derivation_nfbm_objective}

Similarly to the derivation of the \gls{nfdm} objective in \Cref{app:derivation_nfdm_objective}, here we also apply discretization to derive the objective of \gls{nfbm}.

The conditional reverse process (\cref{eq:app_f_rsde}) and it's discretization (\cref{eq:app_f_rsde_discr}) stays the same, except now the conditional reverse process are conditioned on a pair of points $\x = (\x_0, \x_1)$, sampled from a joined data distribution $(\x_0, \x_1) \sim q(\x_0, \x_1)$. The reverse process (\cref{eq:app_r_rsde}) and its discretization (\cref{eq:app_r_rsde_discr}) also have the same structure. However, for \gls{nfbm} the reverse process starts from distribution $\bp(\z_1|\x_1) = p(\z_1|\x_1)$ and at the end samples from $\bp(\x_0|\z_0) = p(\x_0|\z_0)$.

We derive the variational bound on \gls{nll} of \gls{nfbm}:
\begin{align}
    \E&_{q(\x_0)} \left[ - \log \bp_{\th, \ph}(\x_0) \right] \\
    = & \E_{q(\x_0)} \left[ - \log \int \int \bp(\x_0|\bz_0) \bp_{\th, \ph}(\bz_{0:1-\D t}|\bz_1) \bp(\bz_1|\x_1) q(\x_1) d z_{0:1} d x_1 \right] \\
    = & \E_{q(\x_0)} \left[ - \log \E_{\bq(\x_1, \bz_{0:1}|\x_0)} \left[ \bp(\x_0|\bz_0) \frac{\bp_{\th, \ph}(\bz_{0:1-\D t}|\bz_1) \bp(\bz_1|\x_1) q(\x_1)}{\bq_\ph(\bz_{0:1-\D t}|\bz_1,\x_0,\x_1) \bq_\ph(\bz_1|\x_0,\x_1) q(\x_1|\x_0)} \right] \right] \\
    \leq & \E_{q(\x_0,\x_1,\z_{0:1})} \left[ - \log \bp(\x_0|\bz_0) \frac{\bp_{\th, \ph}(\bz_{0:1-\D t}|\bz_1) \bp(\bz_1|\x_1) q(\x_1)}{\bq_\ph(\bz_{0:1-\D t}|\bz_1,\x_0,\x_1) \bq(\bz_1|\x_0,\x_1) q(\x_1|\x_0)} \right] \\
    = & \bLrec + \bLdiff + \bLprior + \textrm{const}, \quad \textrm{where} \\
    & \bLrec = \E_{q(\x_0,\z_0)} \left[ - \log \bp(\x_0|\bz_0) \right], \\
    & \bLdiff = \sum_{t=\frac{1}{T}}^T \E_{\bq_\ph(\x_0, \x_1, \bz_t)} \left[ \KL \big( 
    \bq_\ph( \bz_{t - \D t} | \bz_t, \x_0, \x_1) \|
    \bp_{\th, \ph}( \bz_{t - \D t} | \bz_t)
    \big) \right], \\
    & \bLprior = \E_{q(\x_0, \x_1)} \left[ \KL \big( \bq_\ph(\bz_1|\x_0,\x_1) \| \bp(\bz_1|\x_1) \big) \right], \\
    & \textrm{const} = \E_{q(\x_0)} \left[ \KL \big( q(\x_1|\x_0) \| q(\x_1) \big) \right].
\end{align}

Here, the reconstruction term $\Lrec = \bLrec$ and the prior term $\Lprior = \bLprior$ do not depend on discretization, and the constant term does not depend on model's parameters. For the diffusion term $\bLdiff$ it is easy to see that in the case of \gls{nfbm} it has the same structure as in \gls{nfdm}. Therefore, taking the limit of the number of discretization steps $T$ the diffusion term $\bLdiff$ transforms to:
\begin{align}
    \L_\mathrm{diff} = \E_{u(t) \bq_\ph(\x_0, \x_1, \bz_t)} \left[
    \frac{1}{2 g^2_\ph(t)} \left\|
    \tfb_\ph(\bz_t, t, \x_0, \x_1) - \hf_{\th, \ph}(\bz_t, t)
    \right\|_2^2
    \right],
\end{align}
where $u(t)$ is a uniform distribution over a unit interval.

This gives us a variational bound on the \acrlong{nll} of the \gls{nfbm}, with a very similar structure as the \gls{nfdm} objective. Additionally, for \gls{nfbm} we also may omit optimization of $\Lrec$ and prior $\Lprior$ when $q_\ph(\z_0|\x_0,\x_1) \approx \d (\x - \z_0)$ and $q_\ph(\z_1|\x_0,\x_1) \approx p(\z_1|\x_1)$.

\section{Details of NFDM}
\label{app:nfdm}

\subsection{Motivation of Generalized Forward Process}
\label{app:nfdm_motivation}

In this section, we provide an extended discussion on the motivation behind the learnable \gls{fproc}.

From a theoretical perspective, diffusion models can be viewed as a hierarchical \gls{vae}, where the objective function of diffusion models is the variational bound on model log-likelihood. Therefore, diffusion models aim to match the joint distributions of data $\x$ and latent variables $\z_{0:1}$ (for simplicity, let's consider a discrete-time model with some $T$ time steps). However, as we know from \cite{kingma2014auto, rezende2014stochastic}, the variational objective is equal to:
\begin{align}
    \underbrace{\KL\big( q_\ph(\x, \z_{0:1}) \| p_\th(\x, \z_{0:1}) \big)}_{\textrm{objective}} = 
    \underbrace{\KL\big( q(\x) \| p_\th(\x) \big)}_{\textrm{entropy + likelihood}} + 
    \underbrace{\KL\big( q_\ph(\z_{0:1} | \x) \| p_\th(\z_{0:1} | \x) \big)}_{\textrm{posterior error}}.
\end{align}

Thus, by minimizing the variational objective in diffusion models, we address two aspects: the likelihood of the data and the posterior error. From this perspective, it is quite a natural choice to also optimize the variational distribution $q_\ph(\z_{0:1} | \x)$ to match the model's posterior distribution $p_\th(\z_{0:1} | \x)$.

In other words, by making the \gls{fproc} learnable, we optimize not only the \gls{rproc} to match the \gls{fproc} but also allow the \gls{fproc} to match the \gls{rproc}. This approach simplifies the task of matching the data distribution by the \gls{rproc}.

\subsection{Calculation of the Time Derivative}
\label{app:nfdm_time_dir}

As discussed in \Cref{sec:nfdm_forward}, the definition of \gls{fproc} requires access to the time derivative of the function $F_\ph$. This derivative is crucial for constructing a conditional \gls{ode}~(see \cref{eq:f_ode}):
\begin{align}
    f_\ph(\z_t, t, \x) = \left. \frac{\p F_\ph(\e, t, \x)}{\p t} \right|_{\e=F_\ph^{-1}(\z_t, t, \x)}.
    \label{eq:app_f_ode}
\end{align}

First, let us clarify this notation: to find the \gls{ode} drift term $f_\ph$ at point $\z_t$ and time $t$ conditionally on $\x$, we first apply the inverse function $F_\ph^{-1}$ to identify $\e$, which corresponds to $\z_t$, $t$, and $\x$. Subsequently, we compute the partial derivative of $F_\ph$ with respect to $t$ at $\e$ and $\x$.

For certain parameterizations of $F_\ph$, the time derivative can be determined analytically. For instance, $F_\ph$ associated with conventional diffusion models (see \Cref{app:connections_cases}) have time derivatives available in a closed form. However, in the general case, the time derivative is not readily available.

Fortunately, we can employ automatic differentiation tools like PyTorch \citep{paszke2017automatic} or JAX \citep{jax2018github}. The naive approach to determining the time derivative involves running a backpropagation procedure for each output of $F_\ph$ and reconstructing each coordinate of the time derivative. Unfortunately, this method is not scalable for high-dimensional $F_\ph$.

Instead of backpropagation, which corresponds to a \gls{vjp}, we can use a forward differentiation mode, or \gls{jvp}. By fixing the inputs $\e$ and $\x$ of the function $F_\ph$ and focusing solely on the input $t$, the Jacobian of $F_\ph$ appears as a vector. Consequently, computing a \gls{jvp} with a unit one-dimensional vector precisely produces the time derivative of $F_\ph$.

This approach allows for scalable computation of time derivatives.

\subsection{Calculation of the Conditional Score Function}
\label{app:nfdm_score}

In \Cref{sec:nfdm_forward}, we define the \gls{fproc} as a conditional \gls{sde}~(\cref{eq:f_sde}):
\begin{align}
    d \z_t &= \tff_\ph(\z_t, t, \x) d t + g_\ph(t) d \w, \quad \textrm{where}\\
    \tff_\ph(\z_t, t, \x) &= f_\ph(\z_t, t, \x) + \frac{g_\ph^2(t)}{2} \n_{\z_t} \log q_\ph(\z_t|\x). \nonumber
\end{align}

To construct this \gls{sde}, access to the conditional score function $\n_{\z_t} \log q_\ph(\z_t|\x)$ is required. This section discusses methods for calculating this score function.

First, consider the case where $F_\ph$ is linear in $\e$ (further details on this parameterization are discussed in \Cref{app:nfdm_parameterization}):
\begin{align}
    F_\ph(\e, t, \x) = \mu_\ph(\x, t) + \e \s_\ph(\x, t),
\end{align}
where $\mu_\ph: \R^D \times [0,1] \mapsto \R^D$ and $\s_\ph: \R^d \times [0,1] \mapsto \R^D$, with the product $\e \s_\ph(\x, t)$ being element-wise. Consequently, the marginal probability densities of the \gls{fproc} are Gaussian:
\begin{align}
    q_\ph(\z_t|\x) = \N(\z_t; \mu_\ph(\x, t), \s_\ph^2(\x, t)).
\end{align}
In the case of Gaussian distributions, the score function can be easily found as:
\begin{align}
    \n_{\z_t} \log q_\ph(\z_t|\x) = \frac{\mu_\ph(\x, t) - \x}{\s_\ph^2(\x, t)} = \left. - \frac{\e}{\s_\ph(\x, t)} \right|_{\e=F_\ph^{-1}(\z_t, t, \x)}.
\end{align}

Now, consider the more general case where the distribution $q_\ph(\z_t|\x)$ is defined implicitly through an invertible transformation $F_\ph$ of the random variable $\e \sim q(\e)$ into $\z_t$. To find the score function, we apply the change of variable formula:
\begin{align}
    & \n_{\z_t} \log q_\ph(\z_t|\x) = \n_{\z_t} \left[ \log q(\e) \Big|_{\e = F_\ph^{-1}(\z_t, t, \x)} + \log \left| \frac{\p F^{-1}_\ph(\z_t, t, \x)}{\p \z_t} \right| \right].
\end{align}

The first part of the equation is the log-density of the noise distribution, which is straightforward to calculate. The second part is the log-determinant of the Jacobian matrix of the inverse transformation, $\log |J_F^{-1}|$. If we have access to $\log |J_F^{-1}|$, we can calculate $\log q\ph(\z_t|\x)$ using the change of variable formula and subsequently compute the gradient using automatic differentiation tools such as PyTorch \citep{paszke2017automatic} or JAX \citep{jax2018github}.

The log-determinant $\log |J_F^{-1}|$ can be easily calculated when $F_\ph$ is linear in $\e$, a case already considered above. For nonlinear, low-dimensional $F_\ph$, the Jacobian $J_F^{-1}$ can be computed using $\O(D)$ backpropagations, and the log-determinant subsequently calculated. However, this approach is not scalable for high-dimensional cases. It is challenging to compute the log-determinant $\log |J_F^{-1}|$ in general high-dimensional cases, posing a limitation to the \gls{nfdm} framework as it restricts the parameterization of the \gls{fproc} to functions with accessible log-determinants.

Nevertheless, the \gls{fproc} can be parameterized with functions $F_\ph$ that inherently provide access to the log-determinant $\log |J_F^{-1}|$ by design. For example, normalizing flow architectures \cite{dinh2017density, kingma2018glow} facilitate the construction of \glspl{fproc} with non-Gaussian, learnable distributions $q_\ph(\z_t|\x)$.

\subsection{Stochastic and Deterministic Sampling}
\label{app:nfdm_sampling}

In \Cref{sec:nfdm_reverse}, we define the \gls{rproc} as the following \gls{sde}~(see \cref{eq:r_rsde}):
\begin{align}
    d \z_t = \hf_{\th, \ph}(\z_t, t) d t + g_\ph(t) d \bw, 
    \quad \textrm{where} \quad
    \hf_{\th, \ph}(\z_t, t) = \tfb_\ph \big( \z_t, t, \hx_\th(\z_t, t) \big).
\end{align}

Let us now rewrite this equation, substituting the definition of $\tfb_\ph$ from \cref{eq:f_rsde}:
\begin{align}
    d \z_t = \left[ 
        f_\ph \big( \z_t, t, \hx_\th(\z_t, t) \big) - 
        \frac{g_\ph^2(t)}{2} \n_{\z_t} \log q_\ph \big( \z_t|\hx_\th(\z_t, t) \big) 
    \right] d t + g_\ph(t) d \bw.
    \label{eq:app_r_rsde_full}
\end{align}

This expression represents the full form of the generative \gls{sde}, for which we derive the objective in \Cref{app:derivation_nfdm_objective}. After training, to sample data points from the models, we can simulate this \gls{sde} as outlined in \cref{eq:app_r_rsde_full} and depicted in \Cref{alg:sampling}.

However, as evident from \cref{eq:app_r_rsde_full}, the generative process has a very simple dependence on $g_\ph(t)$, which determines the level of stochasticity. Therefore, in practice, we can adjust the stochasticity level by modifying the function $g_\ph(t)$.

In the extreme case where $g_\ph(t) \equiv 0$, the generative \gls{sde} completely loses its stochasticity and becomes an \gls{ode}:
\begin{align}
    d \z_t = f_\ph \big( \z_t, t, \hx_\th(\z_t, t) \big)  d t.
\end{align}

This \gls{ode} allows for an interpretation of the model as a \gls{cnf} \cite{chen2018neural, grathwohl2018scalable}, consequently enabling deterministic sampling from a pre-trained model and likelihood estimation.

\subsection{Parameterization of the Forward Process}
\label{app:nfdm_parameterization}

The parameterization of the variance function $g_\ph(t)$~(\cref{eq:f_sde}) and the data point predictor $\hx_\th(\z_t, t)$~(\cref{eq:r_rsde}) is straightforward. In our experiments, we parameterize them directly using neural networks. However, parameterizing the transformation $F_\ph$ (\cref{eq:f}) is more complex. \gls{nfdm} requires $F_\ph$ to be differentiable and invertible with respect to $\e$,  as well as provide access to the log-determinant of the Jacobian matrix (see \Cref{app:nfdm_score}). We propose two parameterizations for the \gls{fproc} of \gls{nfdm}.

First, we propose a parameterization of $F_\ph$ that is linear in $\e$:
\begin{align}
    F_\ph(\e, t, \x) &= \mu_\ph(\x, t) + \e \s_\ph(\x, t), \quad \textrm{where}
    \label{eq:app_f_lin_param} \\
    \mu_\ph(\x, t) &= (1 - t) \x + t (1 - t) \bmu_\ph(\x, t), 
    \label{eq:app_f_lin_param_mu} \\
    \s_\ph(\x, t) &= \d^{1-t} \left( \bs_\ph(\x, t) \right)^{t(1-t)}.
    \label{eq:app_f_lin_param_s}
\end{align}
Here, $\bmu_\ph: \R^D \times [0,1] \mapsto \R^D$ and $\bs_\ph: \R^D \times [0,1] \mapsto \R^D_+$, with $\d$ as a constant. This parameterization models $q(\z_t|\x)$ as a conditional Gaussian distribution:
\begin{align}
    q_\ph(\z_t|\x) = \N(\z_t; \mu_\ph(\x, t), \s_\ph^2(\x, t)).
\end{align}

In our experiments, we set $\d^2 = 10^{-4}$. If $\bmu_\ph$ is differentiable and $\bs_\ph$ is differentiable and non-negative, this parameterization meets the requirements on $F_\ph$.

The functions $\mu_\ph(\x, t)$ and $\s_\ph(\x, t)$ are specifically designed such that at $t=0$, they equal $\mu_\ph(\x, 0) = \x$ and $\s_\ph(\x, t) = \d$, and at $t=1$, they become $\mu_\ph(\x, 1) = 0$ and $\s_\ph(\x, t) = I$. At intermediate times $0 < t < 1$, these functions can take arbitrary values, thereby ensuring that $q(\z_0|\x) = \N(\z_0; \x, \d^2 I)$ and $q(\z_1|\x) = \N(\z_1; 0, I)$, with arbitrary Gaussian distributions at intermediate steps. These restrictions are not necessary for $F_\ph$, however, it eliminates the need to optimize the reconstruction loss $\Lrec$ and the prior loss $\Lprior$, as they become independent of the parameters $\ph$ and $\th$ (see \Cref{app:derivation_nfdm_objective}).

Secondly, we propose a more general parameterization:
\begin{align}
    F_\ph(\e, t, \x) &= (1-t) \x + \big( \d + (1-\d)t \big) \bF_\ph(\e, t, (1-t)\x),
    \label{eq:app_f_flow_param}
\end{align}
where $\bF_\ph: \R^D \times [0,1] \times \R^D \mapsto \R^D$ is an invertible and differentiable function that allows access to the log-determinant of the Jacobian matrix. If $\bF_\ph$ is sufficiently flexible (e.g., a neural network with a normalizing flow architecture), it can parameterize non-Gaussian distributions $q_\ph(\z_t|\x)$.

The additive term $(1-t) \x$ provides an inductive bias that shifts the mean of the distribution $q_\ph(\z_t|\x)$ towards $\x$ when $t$ is near $0$. The linear function $\d + (1-\d)t$ scales the distribution from a factor of $\d$ at $t=0$ to $1$ at $t=1$, also imparting an inductive bias for increasing variance over time. The scaling $(1-t)\x$ in the third argument of $\bF_\ph$ ensures that $q_\ph(\z_1|\x)$ at $t=1$ does not depend on $\x$, allowing the reverse process to start from $p(\z_1) = q_\ph(\z_1|\mathbf{0})$ and thereby guaranteeing matching of prior distributions between the forward and reverse processes. Consequently, this eliminates the need for optimizing the prior loss $\Lprior$. However, unlike the previous parameterization that guarantees $q(\z_0|\x) = \N(\z_0; \x, \d^2 I)$, in this case, there are no guarantees on the form of $q(\z_0|\x)$, necessitating the optimization of the reconstruction loss $\Lrec$ with respect to parameters $\ph$ (see \Cref{app:derivation_nfdm_objective}).

Importantly, even the Gaussian parameterization of $F_\ph$ in \cref{eq:app_f_lin_param} offers a significantly more flexible \gls{fproc} compared to prior works. Although linear in $\e$, it may exhibit complex non-linear dependencies on $\x$ and $t$.

Furthermore, we emphasize that the parameterizations of $F_\ph$, $g_\ph(t)$, and $\hx_\th(\z_t, t)$ discussed here are not the only possible ones and may be sub-optimal. For the purposes of this paper, we maintain a simple setup and leave the exploration of more effective parameterizations for future research.

\section{Details of NFBM}
\label{app:nfbm}

\subsection{Parameterization of the Forward Process}
\label{app:nfbm_forward}

To parameterize the \gls{fproc} of \gls{nfbm}, we adopt the Gaussian parameterization discussed for \gls{nfdm} in \Cref{app:nfdm_parameterization}:
\begin{align}
    F_\ph(\e, t, \x_0, \x_1) &= \mu_\ph(\x_0, \x_1, t) + \e \s_\ph(\x_0, \x_1, t), \quad \textrm{where}
    \label{eq:app_nfbm_f_lin_param} \\
    \mu_\ph(\x_0, \x_1, t) &= (1 - t) \x_0 + t \x_1 + t (1 - t) \bmu_\ph(\x_0, \x_1, t), 
    \label{eq:app_nfbm_f_lin_param_mu} \\
    \s_\ph(\x_0, \x_1, t) &= \d \left( \bs_\ph(\x_0, \x_1, t) \right)^{t(1-t)}.
    \label{eq:app_nfbm_f_lin_param_s}
\end{align}
Here, $\bmu_\ph: \R^D \times \R^D \times [0,1] \mapsto \R^D$ and $\bs_\ph: \R^D \times \R^D \times [0,1] \mapsto \R^D_+$, with $\d^2=10^{-4}$ in our experiments. Similar to the \gls{nfdm}, this parameterization models $q(\z_t|\x)$ as a conditional Gaussian distribution:
\begin{align}
    q_\ph(\z_t|\x_0,\x_1) = \N(\z_t; \mu_\ph(\x_0,\x_1, t), \s_\ph^2(\x_0,\x_1, t)).
\end{align}

In contrast to the parameterization of \gls{nfdm}, here $\mu_\ph(\x, t)$ and $\s_\ph(\x, t)$ are designed to ensure that $q(\z_0|\x_0, \x_1) = \N(\z_0; \x_0, \d^2 I)$ and $q(\z_1|\x_0, \x_1) = \N(\z_1; \x_1, \d^2 I)$. At intermediate times, $0 < t < 1$, $q(\z_t|\x_0, \x_1)$ may still be an arbitrary Gaussian distribution. Therefore, with this parameterization of \gls{nfbm}, we do not need to optimize the reconstruction loss $\Lrec$ and the prior loss $\Lprior$, as they are independent of the parameters $\ph$ and $\th$ (see \Cref{app:derivation_nfbm_objective}).

As with \gls{nfdm}, the parameterization of \gls{nfbm} that we provide in this section represents a simple design choice. We believe there are more efficient ways to parameterize the \gls{fproc} of \gls{nfbm}. However, in this paper, we want to focus on presenting the framework itself, rather than dwelling on design choices.

\subsection{Parameterization of the Reverse Process}
\label{app:nfbm_reverse}

In \Cref{sec:nfbm}, we define the \gls{rproc} of \gls{nfbm} by the following \gls{sde}:
\begin{align}
    d \z_t = \hf_{\th, \ph}(\z_t, t) d t + g_\ph(t) d \bw, 
    \quad \textrm{where} \quad
    \hf_{\th, \ph}(\z_t, t) = \tfb_\ph \big( \z_t, t, \hx^{0,1}_\th(\z_t, t) \big).
\end{align}
Here, the function $\hx^{0,1}_\th(\z_t, t)$ predicts both $\x_0$ and $\x_1$. Assuming $q_\ph(\z_0|\x_0,\x_1) \approx \d (\x - \z_0)$ and $q_\ph(\z_1|\x_0,\x_1) \approx p(\z_1|\x_1)$, this formulation leads to the objective (see \Cref{app:nfbm_forward}):
\begin{align}
    \L = \E_{u(t)q(\x_0, \x_1)q_\ph(\z_t \mid \x_0, \x_1)} \left[ \frac{1}{2 g_\ph^2(t)} \big\| \tfb_\ph(\z_t, t, \x_0, \x_1) - \hf_{\th, \ph}(\z_t, t) \big\|_2^2 \right].
\end{align}

As previously mentioned, alternatively, we could define the reverse process as being fully conditional on the initial data point $\x_1$. This leads to the following \gls{sde}:
\begin{align}
    d \z_t = \hf_{\th, \ph}(\z_t, t) d t + g_\ph(t) d \bw, 
    \quad \textrm{where} \quad
    \hf_{\th, \ph}(\z_t, t) = \tfb_\ph \big( \z_t, t, \hx^0_\th(\z_t, t, \x_1), \x_1 \big).
\end{align}
This approach is akin to classifier-free guidance, where each step of the generative process depends on conditioning $\x_1$. Under the same assumptions—that $q_\ph(\z_0|\x_0,\x_1) \approx \d (\x - \z_0)$ and $q_\ph(\z_1|\x_0,\x_1) \approx p(\z_1|\x_1)$—this definition leads to a slightly different objective (the derivation is not provided here but can be easily inferred by following the steps described in \Cref{app:derivation_nfbm_objective} for the previous definition):
\begin{align}
    \L = \E_{u(t)q(\x_0, \x_1)q_\ph(\z_t \mid \x_0, \x_1)} \left[ \frac{1}{2 g_\ph^2(t)} \big\| \tfb_\ph(\z_t, t, \x_0, \x_1) - \hf_{\th, \ph}(\z_t, t, \x_1) \big\|_2^2 \right].
\end{align}

Intuitively, the difference between these two definitions implies that the first \gls{rproc}, at a point $\z_t$ at time $t$, follows the average trajectory of \glspl{fproc} passing through $\z_t$. Conversely, the second \gls{rproc} tracks the average trajectory of \glspl{fproc} that pass through $\z_t$ and terminate at $\z_1$.

The second approach may be beneficial when dealing with joint data distributions $q(\x_0, \x_1)$. However, in this paper, as we work with unpaired distributions $q(\x_0)q(\x_1)$, we have opted for the first approach in \Cref{sec:nfbm}.

\section{NFDM with restrictions}
\label{app:restrictions}

\subsection{Additional Discussion of NFDM with Restrictions}
\label{app:restrictions_discussion}

In Section \ref{sec:restricted}, we discussed training the \gls{nfdm} with restrictions on the \gls{rproc}. This section aims to further explore the training under restrictions and to examine the limitations of this technique.

Typically, there is an infinite number of forward and reverse processes that correspond to each other. So how to choose between them? The standard \gls{nfdm}, as described in \Cref{sec:nfdm}, learns just one such pair of processes. However, the flexibility of \gls{nfdm} opens up the possibility of choosing specific pairs of processes depending on the task.

Suppose we aim to learn a \gls{rproc} with some specific properties. To achieve this, we must impose certain restrictions on the \gls{rproc}. One approach is to encode these restrictions directly in the parameterization of the \gls{rproc}. Alternatively, we can introduce penalties on the \gls{rproc} that would penalize deviations from the desired properties. In both cases, we expect the \gls{fproc} to adapt and align with the \gls{rproc}. Consequently, the \gls{rproc} will match the \gls{fproc} and, as a result, the data distribution, while possessing the desired properties.

In Section \ref{sec:restricted}, we discussed the restriction of the \gls{rproc} to have straight deterministic generative trajectories. However, we may apply the same strategy to ensure arbitrary properties. We believe that the \gls{nfdm} framework may help construct various new generative dynamics for a variety of tasks and domains.

Nevertheless, training the \gls{nfdm} with restrictions has limitations. The primary concern is the feasibility of these restrictions. For instance, imposing overly stringent constraints on the \gls{rproc} (such as fixing its drift term $\hf \equiv 0$) will render the \gls{fproc} incapable of adapting, regardless of its flexibility. Similarly, unattainable penalties, such as those on the squared norm of the drift term $\| \hf_{\th, \ph} \|_2^2$, will lead to biased solutions, as it is impossible to match two processes when $\| \hf_{\th, \ph} \|_2^2 \equiv 0$.

Additionally, even when a restriction is feasible, if the model parameterization is too simplistic, it may not be able to match processes and satisfy the restriction, leading to bias. However, in our experiments, we find that even the Gaussian parameterization of the \gls{fproc} (see \Cref{app:nfdm_parameterization}) is flexible enough in practice to successfully learn various restrictions, such as straight generative trajectories.

\subsection{Details of NFDM with Curvature Penalty}
\label{app:restrictions_ot}

In \Cref{sec:restricted}, we suggest penalizing the curvature of deterministic generative trajectories specifically with the $\Lcrv$~(\cref{eq:l_crv}) penalty.

To calculate the curvature penalty $\Lcrv$~(\cref{eq:l_crv}), we proceed as follows. First, we use the chain rule to rewrite the time derivative:
\begin{align}
    \frac{d \hf_{\th, \ph}(\z_t, t)}{d t} 
    &= \frac{\p \hf_{\th, \ph}(\z_t, t)}{\p \z_t} \frac{\p \z_t}{\p t} + \frac{\p \hf_{\th, \ph}(\z_t, t)}{\p t} \\
    &= \frac{\p \hf_{\th, \ph}(\z_t, t)}{\p \z_t} \hf_{\th, \ph}(\z_t, t) + \frac{\p \hf_{\th, \ph}(\z_t, t)}{\p t}.
    \label{eq:app_l_crv}
\end{align}

The second term in \cref{eq:app_l_crv} is the time derivative of a function. As discussed in \cref{app:nfdm_time_dir}, the time derivatives can be determined as a \acrfull{jvp} with a one-dimensional unit vector. Notably, the first term in \cref{eq:app_l_crv} is also a \gls{jvp}. Therefore, we can combine these two operations. For this purpose, we define a $\R^{D+1}$ dimensional adjoint vector $v$:
\begin{align}
    v = \begin{bmatrix}
    \hf_{\th, \ph}(\z_t, t) \\
    1
    \end{bmatrix}.
\end{align}

Consequently, $\Lcrv$ can be efficiently computed as the \gls{jvp} of $\hf_{\th, \ph}(\z_t, t)$ with the vector $v$.

Additionally, we would like to note a couple of aspects of learning the \gls{nfdm} with a curvature penalty. First, unlike the regular \gls{nfdm}, the curvature penalty necessitates that $F_\ph$ be twice differentiable with respect to time. Second, although we parameterize $\hf_{\th, \ph}$~(\cref{eq:r_rsde}) through $\tfb_\ph$~(\cref{eq:f_rsde}), $\Lcrv$ influences the \gls{fproc} but does not penalize the curvature of the conditional forward trajectories (see illustrations in \Cref{app:additional_nfdm_ot_forward}).

\section{Connections of NFDM with Other Approaches}
\label{app:connections}

In this section we continue discussing connections between \gls{nfdm} and related works.

\subsection{Special Cases of NFDM}
\label{app:connections_cases}

Many existing works define the \gls{fproc} in diffusion models as either fixed or simply parameterized processes. These processes can be considered special cases of \gls{nfdm}. In this section, we review some of these approaches.

\textbf{Soft Diffusion}. \cite{daras2022soft} consider the case of a fixed \glspl{fproc} $q(\z_t|\x) = \N(\z_t; C_t \x, \s_t I)$, which can be reparameterized as $F(\e, t, \x) = C_t \x + \s_t \e$. Such distributions include, for example, combinations of blurring and the injection of Gaussian noise.

\textbf{Star-Shaped Diffusion}. \cite{okhotin2024star} extended conventional diffusion models to include distributions from the exponential family. Although Star-Shaped Diffusion is a discrete-time approach and does not directly correspond to \gls{nfdm}, the latter can work with exponential family distributions through reparameterization functions. For instance, for some continuous one-dimensional distribution $q(\z_t|\x)$, \gls{nfdm} could use $F(\e, t, \x) = a(b(\e), t, \x)$, where $a$ is the inverse \gls{cdf} of $q(\z_t|\x)$, and $b$ is the \gls{cdf} of a unit Gaussian.

\textbf{Variational Diffusion Models}. \cite{kingma2021variational} proposed forward conditional distributions $q_\ph(\z_t|\x)$ as $\N(\z_t; \a_\ph(t) \x, \s_\ph^2(t) I)$ with learnable parameters $\ph$. In the context of \gls{nfdm}, this can be realized by $F_\ph(\e, t, \x) = \a_\ph(t) \x + \s_\ph(t) \e$.

\textbf{LSGM}. \cite{vahdat2021score} suggest an alternative approach for parameterizing the \gls{fproc}, proposing diffusion in a latent space of a \gls{vae}. Therefore, the \gls{fproc} is characterized by a distribution $q_\ph(\z_t|\x) = \N(\z_t; \a_t E_\ph(\x), \s_t^2 I)$, where $E_\ph$ is the encoder of the \gls{vae}. To parameterize the same forward process with \gls{nfdm}, one could use $F_\ph(\e, t, \x) = \a_t E_\ph(\x) + \s_t \e$. To align the reverse process, the reconstruction distribution should be $p(\x|\z_0) = \N(\x; D_\ph(\z_0), \d^2 I)$, where $D_\ph$ is \gls{vae}'s decoder.

\textbf{ShiftDDPM}. \cite{zhang2023shiftddpms} proposed learning a function that shifts the mean and covariance of $q_\phi(\mathbf{z}_t|\mathbf{x})$ by conditional information. In our work, we do not focus on conditional generation. However, we believe that the \gls{nfdm} framework can be readily adapted for conditional generation by incorporating the conditioning variable $\mathbf{c}$ into both the transformation $F_\phi(\e, t, \mathbf{x}, \mathbf{c})$ and the predictor $\hx(\mathbf{z}_t, t, \mathbf{c})$.

\textbf{NDM} and \textbf{DiffEnc}. \cite{bartosh2023neural, nielsen2024diffenc} proposed a more general \gls{fproc}, $q_\ph(\z_t|\x) = \N(\z_t; \a_t f_\ph(\x, t), \s_t^2 I)$, which, unlike LSGM, transforms $\x$ in a time-dependent manner. This \gls{fproc} can also be described in terms of \gls{nfdm} as the special case $F_\ph(\e, t, \x) = \a_t f_\ph(\x, t) + \s_t \e$.

\textbf{Curvature minimization}. \cite{lee2023minimizing} suggested learning the distribution $q_\ph(\z_1|\x)$ of the \gls{fproc} at time step $t=1$ as a conditional Gaussian. They parameterize the forward process via linear interpolation between $\x$ and $\z_1$. Within the \gls{nfdm} framework, we can express this as $F_\varphi(\e, t, \x) = (1 - t) \x + t \big( \mu_\ph(\x) + \s_\ph(\x) \e \big)$. Subsequently, the authors train their model by optimizing the Flow Matching objective in conjunction with minimizing the \gls{kl} divergence between $q_\ph(\z_1|\x)$ and $p(\z)$. This objective corresponds to minimization of diffusion term $\Ldiff$ and prior term $\Lprior$ of \gls{nfdm} objective (see \Cref{app:derivation_nfdm_objective}) for the \gls{fproc} proposed by \cite{lee2023minimizing}.

\subsection{Compatible and Orthogonal Approaches}

Despite the \gls{simfree} nature of the training procedure, diffusion models still necessitate full \gls{rproc} simulations for sample generation, leading to slow and computationally expensive inference. To address this, in an orthogonal line of works alternative sampling methods have been studied, such as deterministic sampling in discrete \cite{song2021denoising} and continuous time \cite{song2021scorebased} and novel numerical solvers \cite{tachibana2021taylor, liu2022pseudo, lu2022dpm, shaul2024bespoke}. Additionally, distillation techniques have been applied to both discrete \cite{salimans2022progressive} and continuous time models \cite{song2023consistency, zheng2023fast, yin2023one} to enhance sampling speed, albeit at the cost of training additional models. Most of these approaches are compatible with \gls{nfdm} and may be adapted for further gains.

The significance of deterministic trajectories in efficient sampling was highlighted by \cite{karras2022elucidating}. At the same time, building on diffusion model concepts, \cite{lipman2023flow, liu2023flow, albergo2023building} introduced \gls{simfree} methods for learning deterministic generative dynamics. Based on these ideas, \cite{liu2023flow} and \cite{liu2022rectified} proposed distillation procedures to straighten deterministic generative trajectories. Since these methods are based on distillation, we consider them orthogonal to \gls{nfdm}.

\subsection{Other Related Works}

\textbf{Schrödinger Bridge models}. Recent studies \cite{zhang2021diffusion, de2021diffusion, shi2024diffusion} have explored generative models based on Schrödinger Bridge theory and finite-time diffusion constructions. Unlike \gls{nfdm}, these models deviate from the \gls{simfree} paradigm by necessitating the full simulation of stochastic processes for training, which significantly increases their computational cost. Moreover, these studies typically aim to address the optimal stochastic control problem, which is outside the scope of our work.

Nevertheless, it is important to highlight that within this research domain, some authors have proposed advanced techniques for parameterizing learnable conditional processes. For instance, in \cite{liu2024generalized}, the authors introduced the use of splines to parameterize the conditional process, thereby constructing a more flexible and efficient model. This parameterization technique can be seen as a specific instantiation of the parameterization procedure proposed by the \gls{nfdm} framework.

\textbf{Minibatch optimal transport}. \cite{pooladian2023multisample} and \cite{tong2023improving} proposed to construct the \gls{fproc} with optimal data-noise couplings to learn straighter generative trajectories. While this method is theoretically justified, it relies on estimating an optimal coupling over minibatches of the entire dataset, which, for large datasets, may become uninformative as to the true optimal transport coupling. In contrast, \gls{nfdm}-OT directly penalizes curvature of the generative trajectories.

\textbf{Stochastic Interpolants}. In this section, we explore the connections between \gls{nfdm} and Stochastic Interpolants, as proposed by \cite{albergo2023stochastic}. Both \gls{nfdm} and Stochastic Interpolants introduce a more general family of \glspl{fproc} through the use of a reparameterization function. Although we acknowledge the relevance of Stochastic Interpolants to our work, there are notable differences between the two approaches.

First, our methodology involves parameterizing the \gls{rproc} by substituting the prediction of $\x$ into the \gls{fproc}, whereas Stochastic Interpolants necessitate learning two separate functions for the \gls{rproc}: the velocity field and the score function.

Second, we present \gls{nfdm} as a framework that enables both the pre-specification and learning of the \gls{fproc}, in contrast to Stochastic Interpolants, which are derived under the assumption of a fixed \gls{fproc}. Consequently, the objectives utilized by Stochastic Interpolants do not support the incorporation of learnable parameters for the \gls{fproc}. Furthermore, these objectives are tailored towards learning the velocity field and the score function, rather than optimizing likelihood, as is the case with \gls{nfdm}.

In their practical applications, Stochastic Interpolants are demonstrated with a simple parameterizations of the \gls{fproc}. They propose a theoretical framework for learning the \gls{fproc} that would result in an \gls{rproc} characterized by dynamical optimal transport. However, this approach is contingent upon solving a high-dimensional min-max optimization problem, for which they do not provide experimental results. Moreover, their work does not clearly articulate how Stochastic Interpolants might be applied to learning other generative dynamics.

In contrast, \gls{nfdm} introduces a more generic method for learning generative dynamics. Moreover, when \gls{nfdm} is learned with restrictions or penalties (see \Cref{sec:restricted}), it remains within a min-min optimization paradigm, which can be addressed more efficiently.

\subsection{Connections of NFDM Objective}
\label{app:connections_objective}

In this section, we delve into the details of the \gls{nfdm}'s objective function (\cref{eq:objective}). Specifically we will consider the diffusion term of the objective $\Ldiff$~(see \Cref{app:derivation_nfdm_objective}).

First, let's unpack the $\Ldiff$ by substituting the definitions of $\tfb_\ph$ and $\hf_{\th, \ph}$:
\begin{align}
    \label{eq:app_objective}
    \Ldiff 
    = \underset{u(t)q_\ph(\x, \z_t)}{\E} & \left[ \frac{1}{2 g_\ph^2(t)} \left\| 
        \tfb_\ph(\z_t, t, \x) - \hf_{\th, \ph}(\z_t, t) 
    \right\|_2^2 \right] \\
    = \underset{u(t)q_\ph(\x, \z_t)}{\E} 
    & \left[ \frac{1}{2 g_\ph^2(t)} \left\|
        \underbrace{ \Big( 
            f_\ph(\z_t, t, \x) - 
            f_\ph \big( \z_t, t, \hx_\th(\z_t, t) \big) 
        \Big) }_{\textrm{Flow Matching term}} +
    \right. \right. \\
    & \quad \left. \left.
        \frac{g_\ph^2(t)}{2} \underbrace{ \Big( 
            \n_{\z_t} \log q_\ph \big( \z_t|\hx_\th(\z_t, t) \big) - 
            \n_{\z_t} \log q_\ph(\z_t|\x) 
        \Big) }_{\textrm{Score Matching term}}
    \right\|_2^2 \right].
    \nonumber
\end{align}

This formulation clearly delineates two components of the objective: the first calculates the difference between the \gls{ode} drift terms, and the second calculates the difference between the score functions. Moreover, this expression highlights the role of $g_\ph(t)$. When $g_\ph(t)$ is small, the forward and reverse processes exhibit smoother trajectories, and the objective is dominated by the first term. Conversely, when $g_\ph(t)$ is large, the processes exhibit more stochastic trajectories, and the objective is dominated by the second term.

Crucially, in the extreme scenarios where $g_\ph(t)$ approaches either $0$ or $\infty$, the diffusion loss $\Ldiff$ corresponds to either a reweighted Flow Matching loss \cite{lipman2023flow} or a reweighted Score Matching loss \cite{vincent2011connection}, respectively:
\begin{align}
    \lim_{g_\ph(t) \rightarrow 0} & \Ldiff 
     = \lim_{g_\ph(t) \rightarrow 0} \E_{u(t)q_\ph(\x, \z_t)} \left[ \frac{1}{2 g_\ph^2(t)} \left\| 
        f_\ph(\z_t, t, \x) - 
        f_\ph \big( \z_t, t, \hx_\th(\z_t, t) \big) 
    \right\|_2^2 \right], \\
    \lim_{g_\ph(t) \rightarrow \infty} & \Ldiff 
     = \lim_{g_\ph(t) \rightarrow \infty} \E_{u(t)q_\ph(\x, \z_t)} \left[ \frac{g_\ph^2(t)}{8} \left\| 
        \n_{\z_t} \log q_\ph \big( \z_t|\hx_\th(\z_t, t) \big) - 
        \n_{\z_t} \log q_\ph(\z_t|\x) 
    \right\|_2^2 \right]
\end{align}

This connection highlights the importance of optimizing the full objective in \cref{eq:app_objective}, which contains both Flow Matching and Score Matching components. Simplifying the objective to include only one component renders it intractable from a variational perspective. Specifically, when $g_\phi(t)$ approaches either $0$ or $\infty$, the objective, a variational bound on likelihood, becomes infinitely large.

There are two primary reasons why many popular diffusion models are trained using just one of these components.

Firstly, many diffusion models employ a simple linear parameterization of the forward process. For these models, both the conditional vector field $f_\ph(\z_t, t, \x)$ and the conditional score function $\n_{\z_t} \log q_\ph(\z_t|\x)$ are merely linear combinations of $\z_t$ and $\x$. As \cite{kingma2024understanding} demonstrated, for such models, Flow Matching and Score Matching can be viewed as reweighted \gls{elbo}. However, for more general non-linear \gls{fproc} in \gls{nfdm}, optimizing Flow Matching and Score Matching objectives is not equivalent.

Secondly, in other models, the Flow Matching and Score Matching objectives are derived under the assumption that the \gls{fproc} is fixed. Without this assumption, we cannot derive either objective in the general case. In \gls{nfdm}, we learn the \gls{fproc} end-to-end with the \gls{rproc}. Thus, reducing the \gls{nfdm}'s objective could lead to the model's collapse, as it is not the correct objective to optimize.

We would also like to emphasize that the \gls{nfdm} framework, including its objective in \cref{eq:app_objective}, represents a generalization of conventional diffusion models, rather than an alternative approach to training. Consequently, if we take an existing \gls{fproc} from conventional diffusion models and incorporate it into the objective $\mathcal{L}$, we will not create a new model or optimization procedure. Instead, this will align with approaches that focus on likelihood-based training of conventional diffusion models, as explored in \cite{song2021maximum, zheng2023improved}.

\gls{nfdm} facilitates the convenient pre-specification of fixed \glspl{fproc}. Hence, it might be utilized to define a fixed forward process and subsequently train it with the Flow Matching objective. However, this is a topic for future research and is beyond the scope of this work.

\section{Implementation Details}
\label{app:implementation}

\begin{table}[!t]
\caption{Training hyper-parameters.}
\label{tab:hyper_parameters}
\centering
\begin{tabular}{lcccc}
\toprule
 & \textbf{CIFAR-10} & \textbf{ImageNet 32} & \textbf{ImageNet 64}  & \textbf{AFHQ 64} \\ \midrule
Channels & 256 & 256 & 192 & 192 \\
Depth & 2 & 3 & 3 & 3 \\
Channels multipliers & 1,2,2,2 & 1,2,2,2 & 1,2,3,4 & 1,2,2,2 \\
Heads & 4 & 4 & 4 & 4 \\
Heads Channels & 64 & 64 & 64 & 64 \\
Attention resolution & 16 & 16,8 & 32,16,8 & 32,16,8 \\
Dropout & 0.0 & 0.0 & 0.0 & 0.0 \\
Effective Batch size & 256 & 1024 & 2048 & 1024 \\
GPUs & 2 & 4 & 16 & 16 \\
Epochs & 1000 & 200 & 250 & 100 \\
Iterations & 391k & 250k & 157k & 102k \\
Learning Rate & 4e-4 & 1e-4 & 1e-4 & 1e-4 \\
Learning Rate Scheduler & Polynomial & Polynomial & Constant & Constant \\
Warmup Steps & 45k & 20k & - & - \\
\bottomrule
\end{tabular}
\end{table}

Our evaluation of \gls{nfdm} and \gls{nfbm} includes tests on synthetic data, CIFAR-10 \cite{krizhevsky2009learning}, two downsampled ImageNet \cite{deng2009imagenet, van2016pixel} datasets, and downsampled AFHQ \cite{choi2020stargan}. To maintain consistency with baselines, we employ horizontal flipping as a data augmentation technique in training models on CIFAR-10 and ImageNet \cite{song2021scorebased, song2021maximum}. For density estimation of discrete data, uniform dequantization is used (see \Cref{app:implementation_dequantization}).

We parameterize $\hx_\th$~(\cref{eq:r_rsde}) in the \gls{rproc} using a 5-layer perceptrons with 512 neurons in each layer for synthetic data and the U-Net architecture from \cite{dhariwal2021diffusion} for images. In all experiments, a 3-layer perceptrons with 64 neurons in each layer is employed to parameterize $g_\ph$~(\cref{eq:f_sde}).

For the parameterization of $F_\ph$(\cref{eq:f}) in \gls{nfdm} as Gaussian, we employ a neural network identical to that used for $\hx_\th$. The only difference is that for $F_\ph$, we double the output of the last layer to parameterize both $\bmu_\ph$~(\cref{eq:app_f_lin_param_mu}) and $\bs_\ph$(\cref{eq:app_f_lin_param_s}) using the same model (see \Cref{app:nfdm_parameterization}).

For parameterizing $F_\ph$~(\cref{eq:nfbm_f}) in \gls{nfdm}, we adopt the same approach but double the input size of the first layer to condition the \gls{fproc} on both $\x_0$ and $\x_1$ (see \Cref{app:nfbm_forward}).

To parameterize $F_\ph$~(\cref{eq:f}) in \gls{nfdm} as non-Gaussian, we apply the Glow \cite{kingma2018glow} architecture tailored to the corresponding dataset to transform $\e$ into $\z_t$. To incorporate conditioning, we use a neural network identical to that of $\hx_\th$ to generate an embedding $u(\x, t)$. We then add this conditioning to the input of each non-linear transformation in the coupling layers.

To ensure the constraints $g_\ph \geq 0$~(\cref{eq:f_sde}) and $\bs_\ph \geq 0$~(\cref{eq:app_f_lin_param_s,eq:app_nfbm_f_lin_param_s}) are met, we apply the softplus function to the outputs of the neural networks.

The models were trained using the Adam optimizer with the following parameters: $\beta_1 = 0.9$, $\beta_2 = 0.999$, a weight decay of $0.0$, and $\epsilon = 10^{-8}$. The training process was facilitated by a polynomial decay learning rate schedule, which includes a warm-up phase for a predefined number of training steps. During this phase, the learning rate is linearly increased from $10^{-8}$ to a peak value. After reaching the peak learning rate, it is then linearly decreased to $10^{-8}$ by the final training step. The specific hyperparameters are detailed in Table \ref{tab:hyper_parameters}. Training was carried out on Tesla V100 GPUs.

\subsection{Dequantization}
\label{app:implementation_dequantization}

For reporting the \gls{nll}, we employ standard uniform dequantization. The NLL is estimated using an importance-weighted average, given by
\begin{align}
    \log \frac{1}{K} \sum_{k=1}^K p_{\theta}(\x + u_k), \quad \text{where} \quad u_k \sim \mathcal{U}(0, 1),
\end{align}
where $\x \in [0, \dots, 255]$.

\subsection{NFBM with Obstacles}
\label{app:implementation_obstacles}

In \Cref{sec:experiments_bridges}, we demonstrate how the concept of learning \gls{nfdm} with the restrictions discussed in \Cref{sec:restricted} can be adapted to \gls{nfbm} for learning generative dynamics that avoid obstacles. For this experiment, we propose training the \gls{nfbm} models with an additional penalty that has higher values at points closer to obstacles. Specifically, we applied the following objective:
\begin{align}
    \L_\textrm{obs} = \E_{u(t)q(\z_t)} [f(\z_t)],
\end{align}
where $f(\z_t)$ is a penalty function. In our experiments, we use the probability density of a mixture of Gaussian distributions, truncated at some low values to $0$. This truncation is necessary to ensure that the model does not push trajectories infinitely away from the obstacle.

In contrast to $\Lcrv$ proposed for \gls{nfdm}-OT, which penalizes the \gls{rproc} influencing the \gls{fproc} as a result, $\L_\textrm{obs}$ directly penalizes the \gls{fproc}, thereby influencing the properties of the \gls{rproc}.

\section{Additional Results}
\label{app:additional}

\subsection{Visualisations of Forward Trajectories of NFDM-OT}
\label{app:additional_nfdm_ot_forward}

\begin{figure}[tp]
    \centering
    \includegraphics[width=0.7\textwidth]{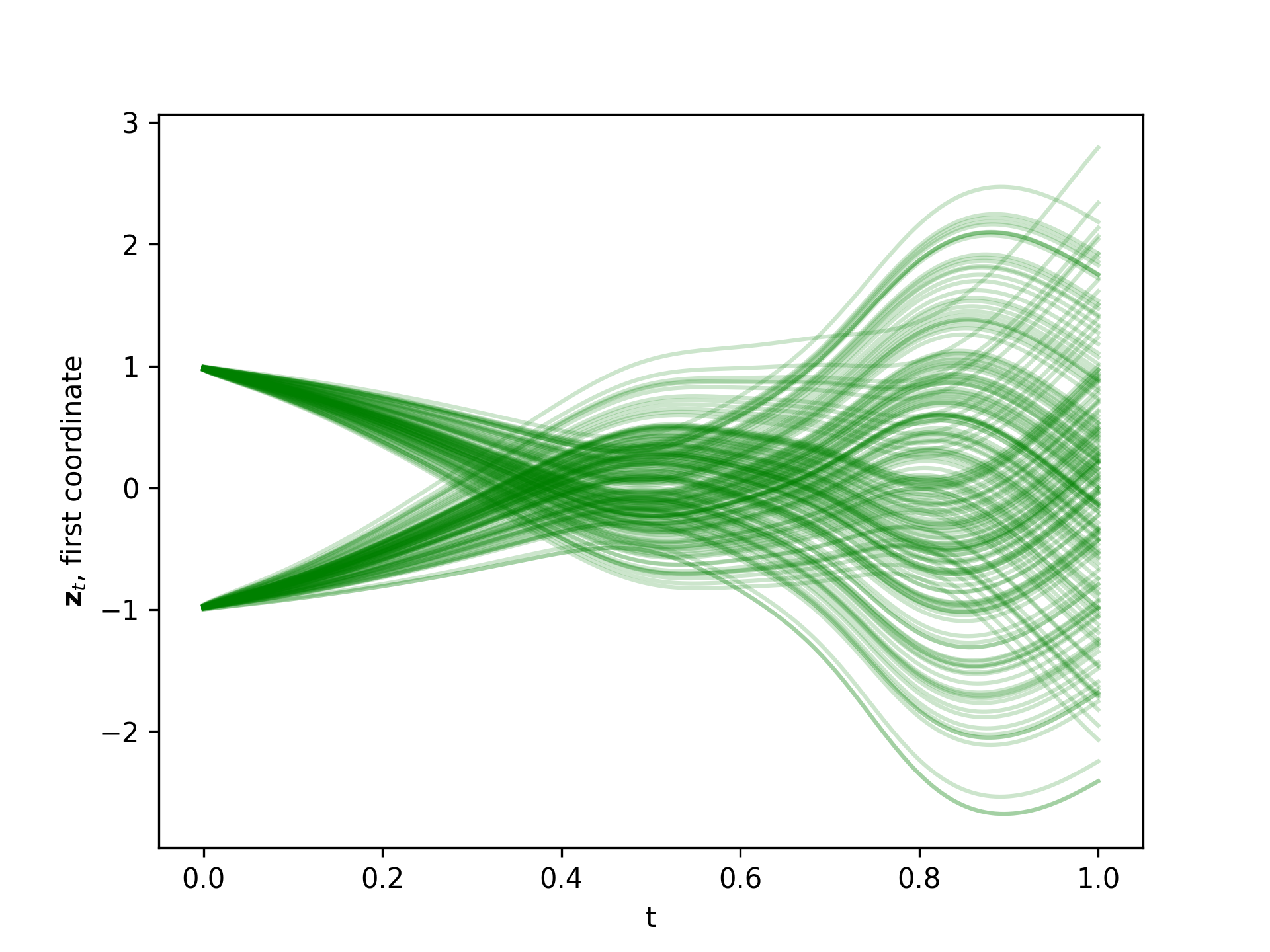}
    \caption{First coordinates of forward deterministic trajectories of \gls{nfdm}-OT started from points $\x=(-1,-1)$ and $\x=(1,1)$.}
    \label{fig:app_nfdm_ot_forward}
\end{figure}

In this section, we provide visualizations of the forward trajectories of \gls{nfdm}-OT, as illustrated in the two-dimensional data distribution experiment shown in \Cref{fig:traj_nfdm_ot}. \Cref{fig:app_nfdm_ot_forward} depicts the deterministic trajectories of \gls{nfdm}-OT. It is easy to see that the forward process has learned some highly curved trajectories. This example highlights several key aspects of \gls{nfdm}-OT and \gls{nfdm} more broadly.

Firstly, this example underscores the point discussed in \Cref{app:restrictions_ot}: while \gls{nfdm}-OT penalizes the curvature of the reverse process, which is parameterized through the forward process (see \Cref{sec:nfdm_reverse}), it does not penalize the curvature of the forward process itself.

Secondly, this example demonstrates the importance of learning nonlinear \glspl{fproc}. In attempting to adapt to the straight-line \gls{rproc}, the \gls{fproc} ends up being highly curved, which in turn aids the \gls{rproc}. This emphasizes the importance of the \gls{nfdm} framework, which facilitates the learning of such nonlinear \glspl{fproc} that may help to learn better generative dynamics.

Finally, we note that we parameterize the \gls{fproc} with a Gaussian distributions featuring learnable mean and variance (see \Cref{app:nfdm_parameterization}). Even this relatively simple parameterization significantly enhances the flexibility of diffusion models compared to conventional diffusion approaches, enabling the learning of complex \glspl{fproc}.

\subsection{Additional Results of NFDM and NFDM-OT on Image Datasets}
\label{app:additional_nfdm_images}

\begin{table*}[!t]
\caption{Comparison of \gls{nfdm} with \gls{nfdm}-OT results on density estimation tasks. We present results in terms of \acrshort{bpd}, lower is better.}
\label{tab:app_nll}
\centering
\begin{tabular}{llcc}
\toprule
Model &  & \textbf{CIFAR10} & \textbf{ImageNet 32}  \\ \cmidrule(r){1-1} \cmidrule(r){3-4}
\gls{nfdm} & & $2.49$ & $3.36$ \\
\cellcolor[HTML]{EFEFEF}\gls{nfdm}-OT & \cellcolor[HTML]{EFEFEF} & \cellcolor[HTML]{EFEFEF}$\mathbf{2.62}$ & \cellcolor[HTML]{EFEFEF}$\mathbf{3.45}$ \\
\bottomrule
\end{tabular}
\end{table*}

\begin{table*}[!t]
\caption{Comparison of \gls{nfdm} with \gls{nfdm}-OT results on few-step generation. We present results in terms of \acrshort{fid}, lower is better.}
\label{tab:app_ot}
\centering
\small
\begin{tabular}{llccccc}
\toprule
 &  & \multicolumn{2}{c}{\textbf{CIFAR-10}} &  & \multicolumn{2}{c}{\textbf{ImageNet 32}} \\
Model &  & \acrshort{nfe} $\downarrow$ & \acrshort{fid} $\downarrow$ &  & \acrshort{nfe} $\downarrow$ & \acrshort{fid} $\downarrow$ \\ \cmidrule(r){1-1} \cmidrule(r){3-4} \cmidrule(r){6-7}
 &  & $4$ & $50.12$ &  & $4$ & $57.60$\\
\multirow{-2}{*}{\gls{nfdm}} &  & $12$ & $21.88$ &  & $12$ & $24.74$\\
\cellcolor[HTML]{EFEFEF} & \cellcolor[HTML]{EFEFEF} & \cellcolor[HTML]{EFEFEF}$4$ & \cellcolor[HTML]{EFEFEF}$7.76$ & \cellcolor[HTML]{EFEFEF} & \cellcolor[HTML]{EFEFEF}$4$ & \cellcolor[HTML]{EFEFEF}$6.13$ \\
\cellcolor[HTML]{EFEFEF}\multirow{-2}{*}{\gls{nfdm}-OT} & \cellcolor[HTML]{EFEFEF} & \cellcolor[HTML]{EFEFEF}$12$ & \cellcolor[HTML]{EFEFEF}$5.20$ & \cellcolor[HTML]{EFEFEF} & \cellcolor[HTML]{EFEFEF}$12$ & \cellcolor[HTML]{EFEFEF}$4.11$\\
\bottomrule
\end{tabular}
\end{table*}

In this section we provide additional results, that directly compare \gls{nfdm} and \gls{nfdm}-OT. This results extend \Cref{tab:nll,tab:ot}. As observed, NFDM without any additional restrictions provides a better log-likelihood estimation compared to NFDM-OT. However, as expected, NFDM exhibits much worse sample quality compared to NFDM-OT in a few steps generation. We attribute this property to the higher curvature of NFDM’s generative trajectories.

\subsection{Generated Samples}
\label{app:additional_samples}

In this section, we present generated samples from the \gls{nfdm}, \gls{nfdm}-OT, and \gls{nfbm} models, trained on various datasets.

\begin{figure}[tp]
    \centering
    \parbox[b]{.32\textwidth}{
        \begin{subfigure}[b]{\linewidth}
            \includegraphics[width=\textwidth]{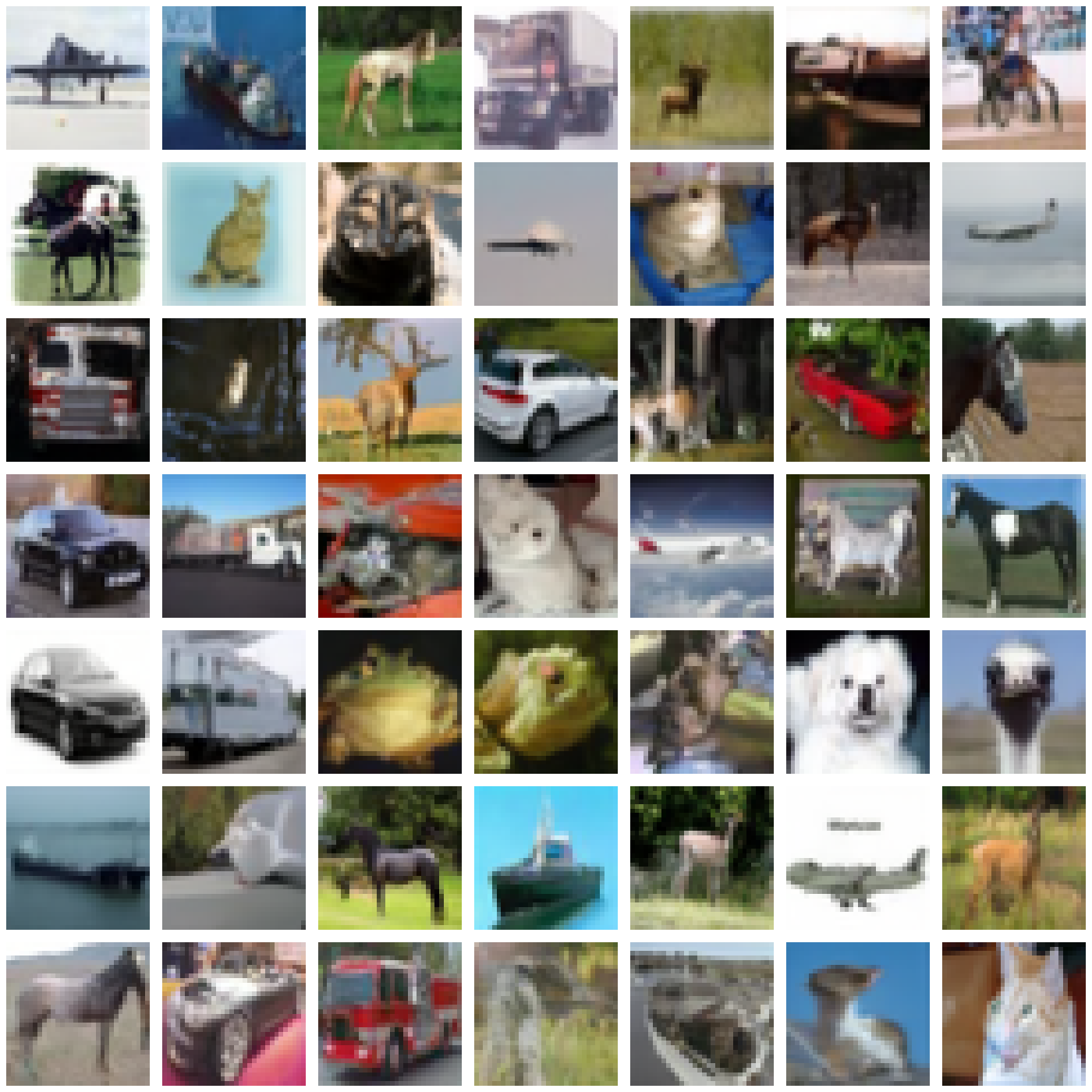}
            \caption{CIFAR-10}
        \end{subfigure}
    }
    \hfill
    \parbox[b]{.32\textwidth}{
        \begin{subfigure}[b]{\linewidth}
            \includegraphics[width=\textwidth]{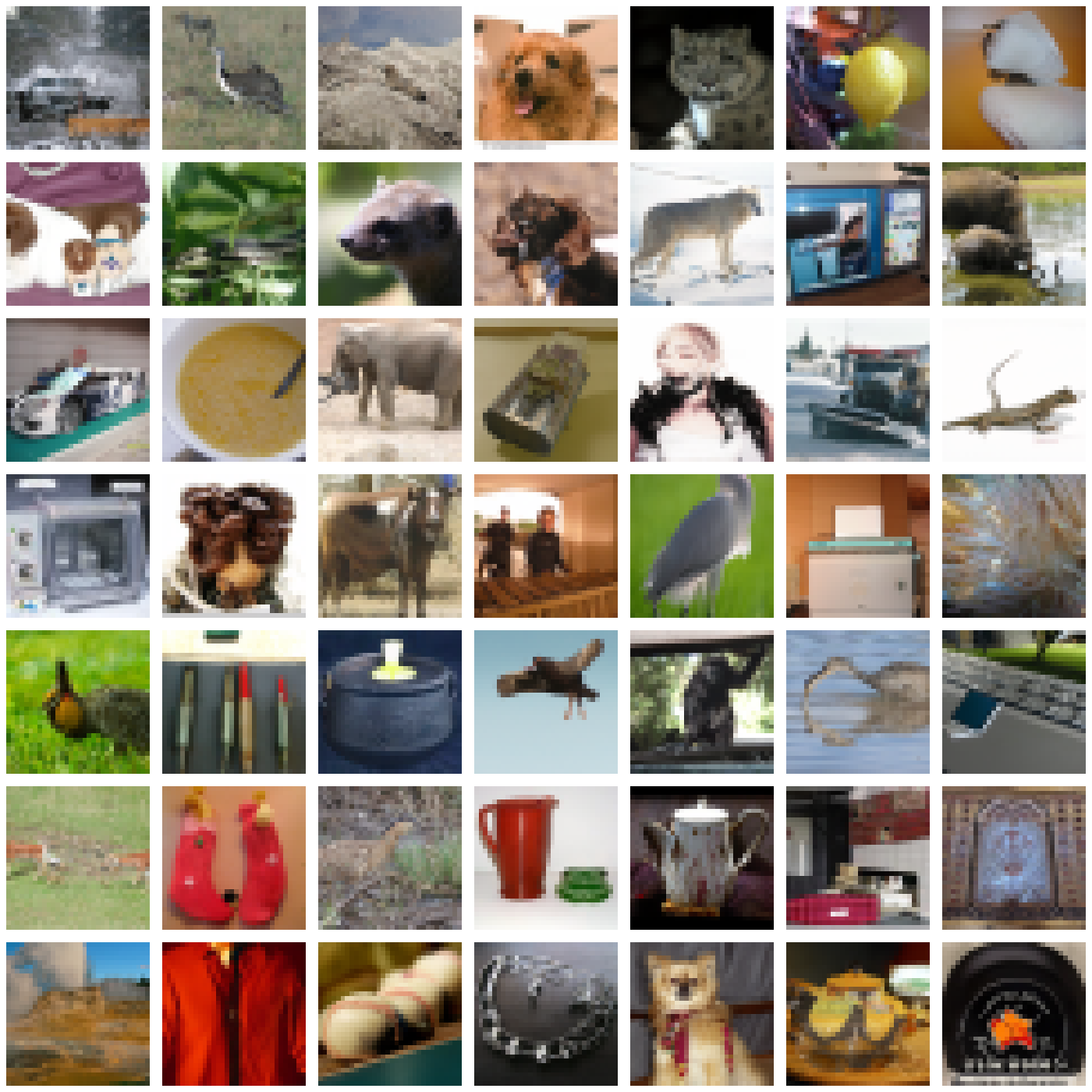}
            \caption{ImageNet 32}
        \end{subfigure}
    }
    \hfill
    \parbox[b]{.32\textwidth}{
        \begin{subfigure}[b]{\linewidth}
            \includegraphics[width=\textwidth]{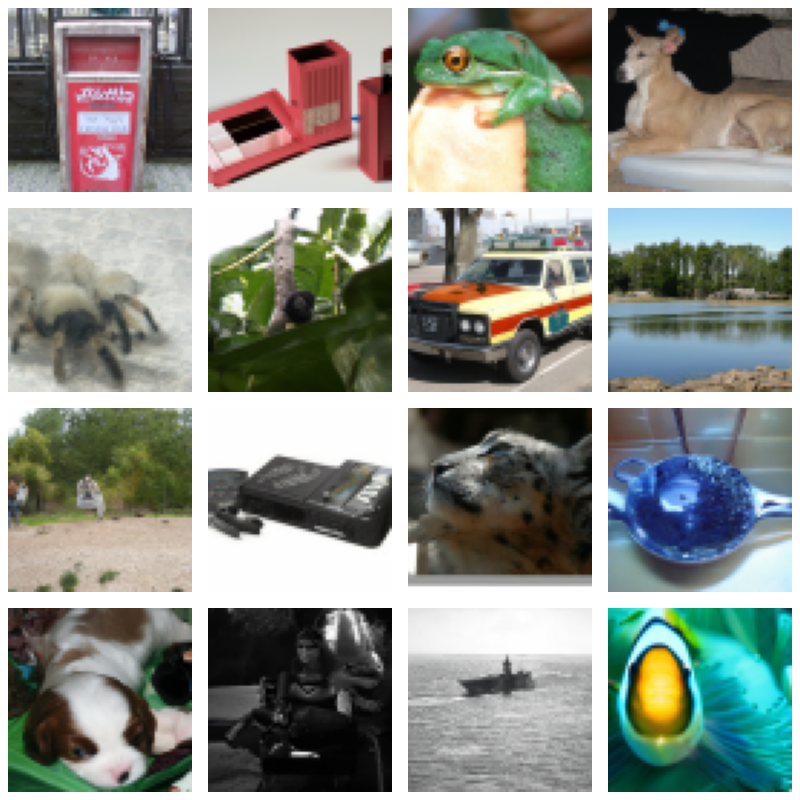}
            \caption{ImageNet 64}
        \end{subfigure}
    }
    \caption{Generated samples from \gls{nfdm} trained on various datasets.}
    \label{fig:app_nfbm_images}
\end{figure}

\textbf{\gls{nfdm}}: \Cref{fig:app_nfbm_images} shows samples from the \gls{nfdm} model with Gaussian $q_\ph(\z_t|\x)$ (see \Cref{app:nfdm_parameterization}). We provide samples from \gls{nfdm} trained on the CIFAR-10, ImageNet 32, and ImageNet 64 datasets.

\begin{figure}[tp]
    \centering
    \includegraphics[width=\textwidth]{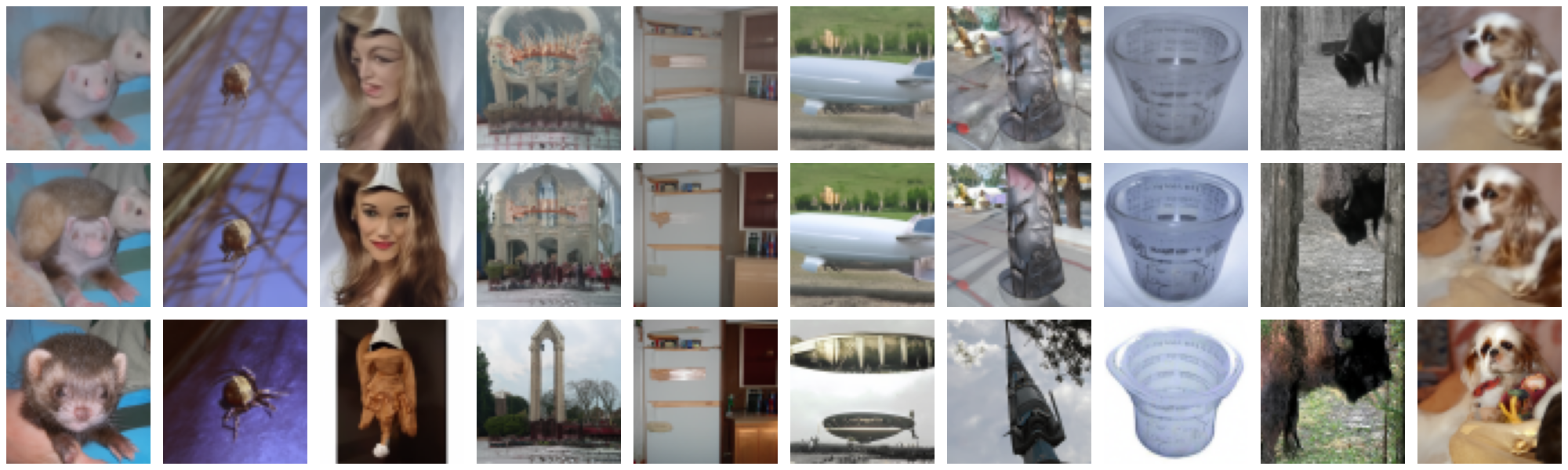}
    \caption{Generated samples from \gls{nfdm}-OT trained on ImageNet 64. \emph{Top:} $\text{\gls{nfe}}=2$; \emph{Middle:} $\text{\gls{nfe}}=4$; \emph{Bottom:} $\text{\gls{nfe}}=12$.}
    \label{fig:app_nfdm_ot_images}
\end{figure}

\textbf{\gls{nfdm}-OT}. \Cref{fig:app_nfdm_ot_images} displays samples from \gls{nfdm} trained on ImageNet 64 dataset. We generate these samples by simulating the \gls{ode} across $12$, $4$, and $2$ steps.

\begin{figure}[tp]
    \centering
    \includegraphics[width=\textwidth]{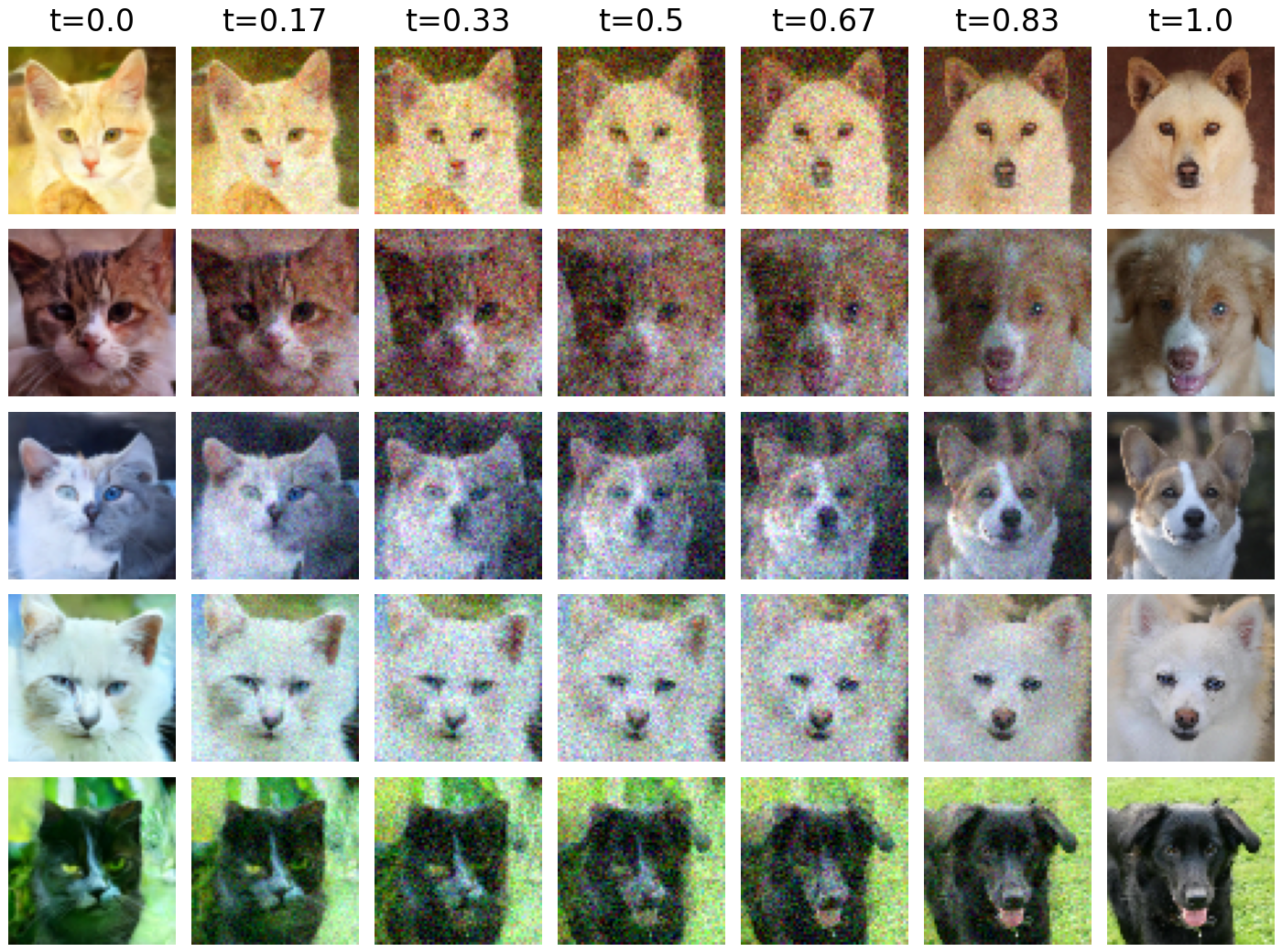}
    \caption{Generative trajectories from \gls{nfbm} trained on AFHQ 64.}
    \label{fig:app_nfbm_afhq}
\end{figure}

\textbf{\gls{nfbm}}. \Cref{fig:app_nfbm_afhq} demonstrates the generative trajectories of the \gls{nfbm} model trained on the AFHQ 64 dataset. This model starts from an image of a dog at time step $t=1$ and simulate the \gls{sde} to time step $t=0$.

\newpage
\section*{NeurIPS Paper Checklist}

\begin{enumerate}

\item {\bf Claims}
    \item[] Question: Do the main claims made in the abstract and introduction accurately reflect the paper's contributions and scope?
    \item[] Answer: \answerYes{} 
    \item[] Justification: Our paper follows the structure of the abstract and introduction directly.

\item {\bf Limitations}
    \item[] Question: Does the paper discuss the limitations of the work performed by the authors?
    \item[] Answer: \answerYes{} 
    \item[] Justification: We discuss limitations in general in \Cref{sec:conclusion} and in \Cref{app:nfdm_score} in more detail.

\item {\bf Theory Assumptions and Proofs}
    \item[] Question: For each theoretical result, does the paper provide the full set of assumptions and a complete (and correct) proof?
    \item[] Answer: \answerYes{} 
    \item[] Justification: We provide proofs in \Cref{app:derivations}.

    \item {\bf Experimental Result Reproducibility}
    \item[] Question: Does the paper fully disclose all the information needed to reproduce the main experimental results of the paper to the extent that it affects the main claims and/or conclusions of the paper (regardless of whether the code and data are provided or not)?
    \item[] Answer: \answerYes{} 
    \item[] Justification: We provide detailed description of the model in \Cref{sec:nfdm,app:nfbm}.

\item {\bf Open access to data and code}
    \item[] Question: Does the paper provide open access to the data and code, with sufficient instructions to faithfully reproduce the main experimental results, as described in supplemental material?
    \item[] Answer: \answerNo{} 
    \item[] Justification: No new data has been generated, and the code will be made available late.

\item {\bf Experimental Setting/Details}
    \item[] Question: Does the paper specify all the training and test details (e.g., data splits, hyperparameters, how they were chosen, type of optimizer, etc.) necessary to understand the results?
    \item[] Answer: \answerYes{} 
    \item[] Justification: The data configuration follows from previous work, while most of the experimental details are provided in \Cref{app:implementation}.

\item {\bf Experiment Statistical Significance}
    \item[] Question: Does the paper report error bars suitably and correctly defined or other appropriate information about the statistical significance of the experiments?
    \item[] Answer: \answerNo{} 
    \item[] Justification: Computation of error bars is vary computationally demanding. We follow standard practice in the literature and report the average result.

\item {\bf Experiments Compute Resources}
    \item[] Question: For each experiment, does the paper provide sufficient information on the computer resources (type of compute workers, memory, time of execution) needed to reproduce the experiments?
    \item[] Answer: \answerYes{} 
    \item[] Justification: We provide compute resources details in \Cref{app:implementation}.
    
\item {\bf Code Of Ethics}
    \item[] Question: Does the research conducted in the paper conform, in every respect, with the NeurIPS Code of Ethics \url{https://neurips.cc/public/EthicsGuidelines}?
    \item[] Answer: \answerYes{} 
    \item[] Justification:  We read and checked the code of ethics.

\item {\bf Broader Impacts}
    \item[] Question: Does the paper discuss both potential positive societal impacts and negative societal impacts of the work performed?
    \item[] Answer: \answerNA{} 
    \item[] Justification: We don't believe this work has any societal impact.
    
\item {\bf Safeguards}
    \item[] Question: Does the paper describe safeguards that have been put in place for responsible release of data or models that have a high risk for misuse (e.g., pretrained language models, image generators, or scraped datasets)?
    \item[] Answer: \answerNA{} 
    \item[] Justification: The paper does not pose such risks.

\item {\bf Licenses for existing assets}
    \item[] Question: Are the creators or original owners of assets (e.g., code, data, models), used in the paper, properly credited and are the license and terms of use explicitly mentioned and properly respected?
    \item[] Answer: \answerYes{} 
    \item[] Justification: All data and software used is properly credited and licenses are respected.

\item {\bf New Assets}
    \item[] Question: Are new assets introduced in the paper well documented and is the documentation provided alongside the assets?
    \item[] Answer:  \answerNA{} 
    \item[] Justification: No assets are introduced in the paper.

\item {\bf Crowdsourcing and Research with Human Subjects}
    \item[] Question: For crowdsourcing experiments and research with human subjects, does the paper include the full text of instructions given to participants and screenshots, if applicable, as well as details about compensation (if any)? 
    \item[] Answer: \answerNA{} 
    \item[] Justification: The paper did not require crowdsourcing.

\item {\bf Institutional Review Board (IRB) Approvals or Equivalent for Research with Human Subjects}
    \item[] Question: Does the paper describe potential risks incurred by study participants, whether such risks were disclosed to the subjects, and whether Institutional Review Board (IRB) approvals (or an equivalent approval/review based on the requirements of your country or institution) were obtained?
    \item[] Answer: \answerNA{} 
    \item[] Justification: No study participants were involved.

\end{enumerate}

\end{document}